\documentclass[10pt,letterpaper]{article}
\PassOptionsToPackage{table}{xcolor}
\usepackage[textwidth=5.5in,textheight=9in,top=1in]{geometry}
\usepackage[T1]{fontenc}
\usepackage{times}
\usepackage{letltxmacro}
\usepackage{newunicodechar}
\newunicodechar{–}{\textendash}
\usepackage[authoryear,round]{natbib}
\setcitestyle{citesep={;},aysep={,},yysep={;}}

\makeatletter
\renewcommand{\maketitle}{\begingroup
  \raggedright
  {\LARGE\scshape \@title\par}
  \vspace{10pt}
  {\centering\normalsize \@author\par}
  \endgroup
  \vspace{8pt}}
\renewcommand{\section}{\@startsection{section}{1}{\z@}{-2ex plus -.5ex minus -.2ex}{1.5ex plus .3ex minus .2ex}{\large\scshape\raggedright}}
\renewcommand{\subsection}{\@startsection{subsection}{2}{\z@}{-1.8ex plus -.5ex minus -.2ex}{.8ex plus .2ex}{\normalsize\scshape\raggedright}}
\renewcommand{\subsubsection}{\@startsection{subsubsection}{3}{\z@}{-1.5ex plus -.5ex minus -.2ex}{.5ex plus .2ex}{\normalsize\scshape\raggedright}}
\makeatother
\renewenvironment{abstract}{\begin{center}\large\scshape Abstract\end{center}
  \begin{quote}\fontsize{10}{11}\selectfont}{\end{quote}}

\LetLtxMacro{\vivaworldvspace}{\vspace}
\ExplSyntaxOn
\cs_new_protected:Npn \vivaworld_vspace_star:n #1
  { \vivaworldvspace * {#1} }
\RenewDocumentCommand \vspace { s m }
  {
    \tl_set:Nn \l_tmpa_tl {#2}
    \tl_replace_all:Nnn \l_tmpa_tl {px} {pt}
    \IfBooleanTF {#1}
      { \exp_args:NV \vivaworld_vspace_star:n \l_tmpa_tl }
      { \exp_args:NV \vivaworldvspace \l_tmpa_tl }
  }
\ExplSyntaxOff

\usepackage{amsmath,amsfonts,bm}

\usepackage{hyperref}
\usepackage{orcidlink}
\newcommand{\authorid}[1]{\,{\hypersetup{hidelinks}\orcidlink{#1}}\,\textsuperscript{1}}
\usepackage{url}
\usepackage{caption}
\usepackage{booktabs}
\usepackage[table]{xcolor}
\definecolor{Palette4}{RGB}{220,235,247}
\definecolor{ScoreSecond}{RGB}{225,242,225}

\usepackage{graphicx}
\usepackage{wrapfig}
\usepackage{needspace}
\definecolor{Palette5}{RGB}{198,219,239}

\title{VivaWorld: Long-Horizon Video World Model with Compositional 3D Priors}

\author{\textbf{Ying Jiang}\authorid{0000-0002-7026-0073} \quad
\textbf{Donglai Xiang}\authorid{0000-0002-6487-1935} \quad
\textbf{Anita Hu}\authorid{0009-0004-2275-4580} \quad
\textbf{Masha Shugrina}\authorid{0000-0002-7583-6772}\\[5pt]
\textsuperscript{1}NVIDIA}
\date{}

\begin{document}

\maketitle

\begin{figure}[!h]
\centering
\includegraphics[width=\textwidth,height=0.40\textheight,keepaspectratio]{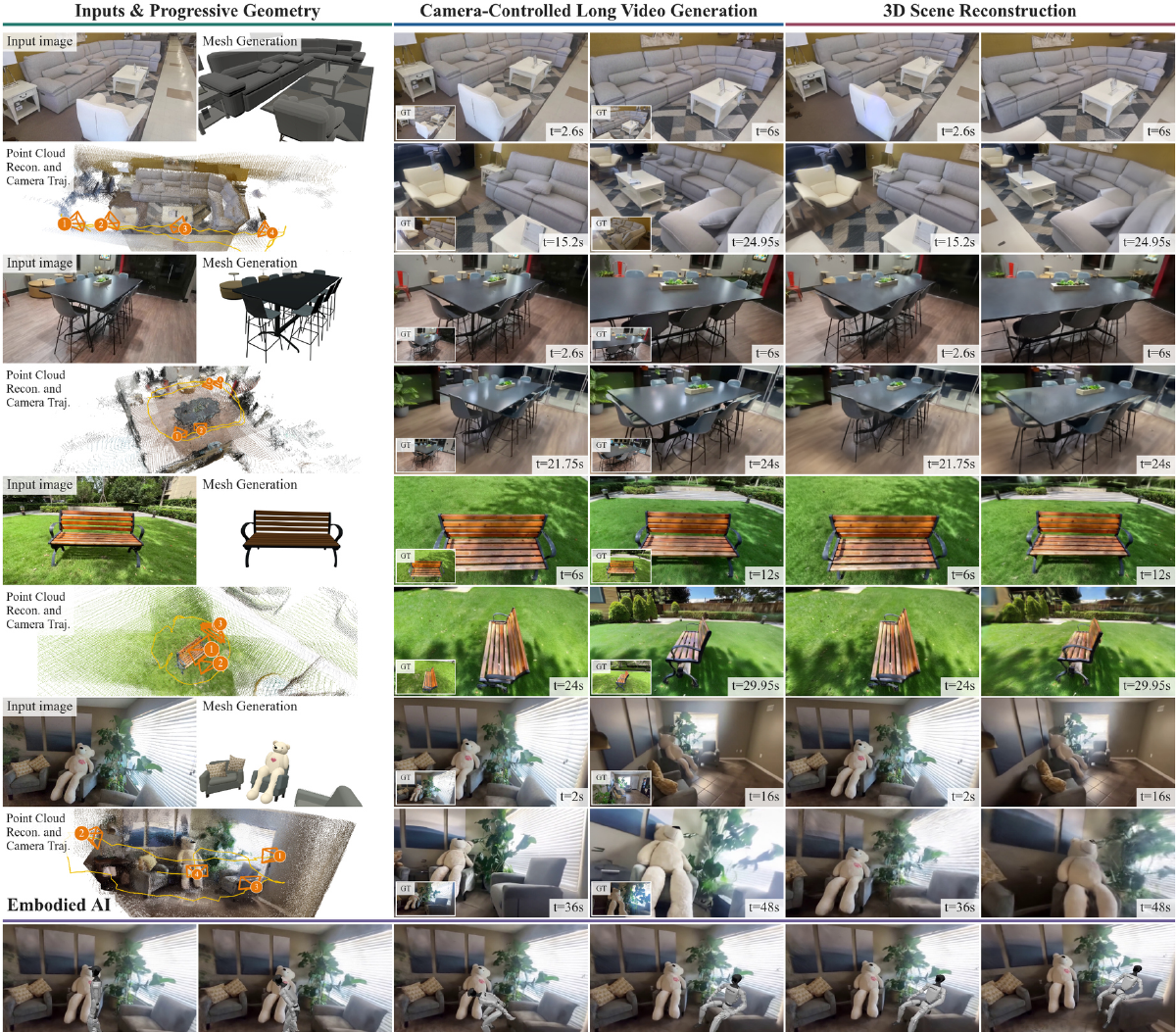}
\caption{VivaWorld is a novel framework that generates long, geometrically consistent, camera-controlled videos from a single image, enabling further compositional scene reconstruction with disentangled foreground objects and downstream embodied AI robot learning applications.}
\label{fig:teaser}
\end{figure}

\begin{abstract}
Long-horizon video generation is key to interactive world modeling, physical intelligence, and spatial reasoning, yet maintaining geometric consistency and accurate camera control remains challenging without 3D grounding. We introduce \textbf{VivaWorld}, a framework for generating geometry- and view-consistent long videos from an image. We first construct a compositional hybrid 3D representation that combines reconstructed point clouds, which preserve observed scene geometry and appearance, with generated textured meshes, which provide complete object geometry across $360^\circ$ novel views. We also introduce a connected-region filter removing inconsistent floating components for robust cross-view reconstruction and propose a hybrid camera embedding that combines global and object-anchored Pl\"{u}cker rays to capture both scene-level camera motion and camera-object relationships, and inject it together with the visual guidance into a video generation model via a control adapter. Extensive experiments show  \textbf{VivaWorld} achieves state-of-the-art visual quality and camera control, while supporting controllable novel-view synthesis, long-horizon scene exploration, and physical AI.
\end{abstract}

\clearpage

\section{Introduction}

Video generation aims to synthesize realistic and spatial-temporally coherent videos from images or text prompts \citep{singer2022make}. Beyond conventional image- or text-conditioned generation, recent video world models are capable of predicting future visual observations conditioned on actions, including camera motions, object trajectories, or agent actions \citep{shen2026lyra, yang2026neoverse, chen2026anyrecon, zheng2026versecrafter}. While early methods primarily focused on generating short video clips conditioned on actions, recent autoregressive video world models \citep{shen2026lyra, chen2026longlive} have enabled the synthesis of substantially longer and more controllable videos with richer motions and dynamics. Although individual frames may appear photorealistic, long-horizon video generation remains challenging, as the lack of explicit 3D supervision results in spatial and temporal inconsistencies, including camera drift, geometry distortion, and object deformation, thereby limiting accurate 3D reconstruction and downstream 3D-aware applications.

To enable spatially and temporally consistent controllable video generation, existing approaches leverage explicit 3D representations~\citep{zheng2026versecrafter, chen2026anyrecon, yang2026neoverse} to reconstruct scenes and render geometrical guidance as 3D memory under target camera viewpoints. Alternatively, recent methods learn implicit geometry priors from 3D latent representations or historical video frames~\citep{huang2026gen3r, shen2026lyra}.
Beyond spatial guidance, recent works \citep{agarwal2026cosmos, yang2026neoverse, shen2026lyra} have also explored more robust camera representations for temporal consistency. However, existing spatial guidance only makes use of 3D information, e.g., point clouds, extracted from the limited input views, providing weak priors for regions unobservable from those views. Moreover, existing camera embeddings are defined in a global coordinate, making them prone to accumulated errors and long-horizon drift. These limitations raise a question: how can we better leverage 3D grounding to enable consistent long video generation?

We present \textbf{VivaWorld}, a framework for generating geometry- and view-consistent 3D-grounded long videos from an image. VivaWorld unifies reconstruction and generative priors through a compositional hybrid representation, providing stronger 3D grounding for long-horizon video generation. Specifically, we leverage feed-forward 3D reconstruction and off-the-shelf 3D generators to reconstruct background point clouds and generate textured foreground meshes. The point clouds preserve observed scene geometry and appearance, while the meshes provide complete object geometry across $360^\circ$ novel viewpoints. Aggregating point clouds reconstructed from multiple views may introduce inconsistent floating artifacts, so we propose a mesh-seeded connected-region filter that removes disconnected artifacts while preserving valid scene structures. To achieve accurate camera control and mitigate long-horizon drift, we further introduce a compositional camera embedding that combines global Pl\"{u}cker rays with object-anchored rays at fixed registered meshes to explicitly capture camera–object relationships. This design reflects how real-world cameras typically move relative to objects and scenes, such as orbiting around a person or panning across a landscape. By jointly encoding the global camera trajectory and local camera-object offsets, our representation captures both scene-level camera motion and object-relative viewpoint changes. We then introduce a compositional control adapter to inject the hybrid RGB-D controls and compositional camera embeddings into a pretrained video diffusion model, enabling the generation of long-horizon videos with consistent geometry, viewpoints, and camera motion. Extensive experiments demonstrate state-of-the-art performance in visual quality and camera control, while highlighting the versatility of VivaWorld across controllable novel-view synthesis, long-horizon scene exploration, and generative 3D reconstruction. To summarize, our contributions are:

\begin{itemize}
    \item We propose \textbf{VivaWorld}, a novel method for generating view- and geometry-consistent 3D-grounded videos from a single image using a compositional hybrid representation of textured object meshes and background point clouds. We integrate a mesh-seeded filtering strategy into 3D point-cloud reconstruction for improved cross-view robustness.

     \item We introduce a compositional camera embedding that combines global and object-anchored representations, enabling stable and accurate camera control by capturing both global camera poses and camera–object relationships.
     \item Extensive experiments demonstrate our method outperforms state-of-the-art approaches in generating accurate, geometry- and view-consistent long videos with precise camera control from a single image, while enabling high-quality compositional 3D reconstruction.
\end{itemize}

\section{Related Work}
\vspace{-5px}
\subsection{Video World Models}
\vspace{-5px}
Recent advances in diffusion models~\citep{rombach2022high} and transformer architectures~\citep{peebles2023scalable} have sparked significant interest in video world models, which aim to predict future states from sparse multimodal inputs, such as actions, text, images, and other observations~\citep{wang2026latent, shen2026lyra, yang2026neoverse, agarwal2026cosmos, yang2025longlive, huang2025voyager, zheng2026versecrafter}. Latent video world models, such as V-JEPA~\citep{assran2025v}, predict future latent representations rather than pixels, enabling video understanding, reasoning, and planning. In contrast, generative world models directly synthesize future video frames from multimodal inputs~\citep{wan2025wan, jiang2025vace, wiedemer2025video}, enabling high-fidelity and controllable generation. However, they often lack explicit 3D supervision, limiting spatial and temporal consistency under complex or long-range camera motion. To address this, existing approaches leverage explicit 3D representations, such as point clouds~\citep{cao2025uni3c, zheng2026versecrafter, chen2026anyrecon, wu2506video, yu2024viewcrafter, yu2025trajectorycrafter}, point tracking \citep{bian2025gs} and 3D Gaussian Splatting~\citep{wu2025difix3d+, yang2026neoverse, de2026artifixer}, to reconstruct scenes and render geometry-aware guidance under target camera viewpoints, providing 3D memory for video generation. Other methods explore historical video frames as complementary guidance~\citep{shen2026lyra} or learn implicit 3D latent representations that encode scene geometry without explicit reconstruction~\citep{huang2026gen3r}. However, existing 3D priors primarily cover regions observed in the input views, leaving unseen regions without reliable geometric guidance and thus with limited 3D consistency. Our goal is to integrate 3D generative priors to cover both observed and unseen regions, enabling spatially and temporally consistent generation.

\vspace{-5px}
\subsection{Camera Control}
\vspace{-5px}
To enable accurate camera control, mainstream approaches condition video generation models on either ray maps or explicit camera embeddings. Ray-based methods typically embed a per-pixel 6D representation consisting of ray origins and directions~\citep{popov2025camctrl3d} or Plücker coordinates~\citep{shen2026lyra, yang2026neoverse, he2024cameractrl, wang2024cpa, he2025cameractrl}. However, dense ray maps rely on a world coordinate, potentially limiting generalization~\citep{li2026cameras}, prompting recent works to explore relative ray embeddings~\citep{zhang2026unified, yin2026raype}. Alternative approaches~\citep{agarwal2026cosmos, popov2025camctrl3d, shin2026trimotion} directly embed camera poses or relative transformations (e.g. to the first or the previous frame) as input. However, none of these representations explicitly model camera–object relationships, causing drift to accumulate over long sequences. Moreover, while point-cloud guidance helps capture camera motion, it may miss coupled drift when the camera and objects shift together, since their pixel-space appearance remains similar. This misalignment propagates to subsequent frames, yielding incorrect camera motion and degraded geometric consistency. Our object-anchored camera embedding anchors camera motion to scene objects to explicitly model their relationships, while hybrid mesh and point-cloud guidance mitigates both camera and object drift, enabling accurate camera control and geometrically consistent long-video generation.

\vspace{-5px}
\subsection{3D Generation and Reconstruction}
\vspace{-5px}
Prior work reconstructs or generates 3D shapes using optimization-based or feed-forward methods across representations such as point clouds, meshes, 3DGS, and NeRFs~\citep{go2025vist3a, schmid2026genrecon, held2026triangle, xia2026sage, bahmani2026lyra, mildenhall2021nerf, kerbl20233d}. While classical per-instance methods, including Structure-from-Motion (SfM)~\citep{schonberger2016structure}, Multi-View Stereo (MVS)~\citep{goesele2006multi}, and optimization-based reconstruction~\citep{mildenhall2021nerf,kerbl20233d} achieve high reconstruction quality, they incur substantial computational costs.
In contrast, recent feed-forward 3D foundation models~\citep{wang2025vggt, wang2026pi, wang2026vggt, lin2025depth, ren2026tokengs, lu2025infinicube, zhao2026rayzer, ren2024scube, wang2024dust3r, wang2025continuous} efficiently infer scene geometry in a single pass, making them well-suited for providing 3D guidance to video generation models. 3D reconstruction and generation methods have different focuses. Reconstruction~\citep{wang2025vggt, wang2026pi, wang2026vggt} provides accurate geometry only in observed regions but offer limited priors for unseen regions; existing work on video generation have explored 3D grounding using this type of priors more. Conversely, 3D generation models~\citep{lai2025hunyuan3d, chen2026sam} provide complete shape priors for arbitrary viewpoints but may not faithfully align with observations. We use the high fidelity of reconstruction and the completeness of generation as complementary hybrid 3D priors for long video generation.

\begin{figure*}[t]
\vspace{-5px}
    \centering
    \includegraphics[width=\textwidth]{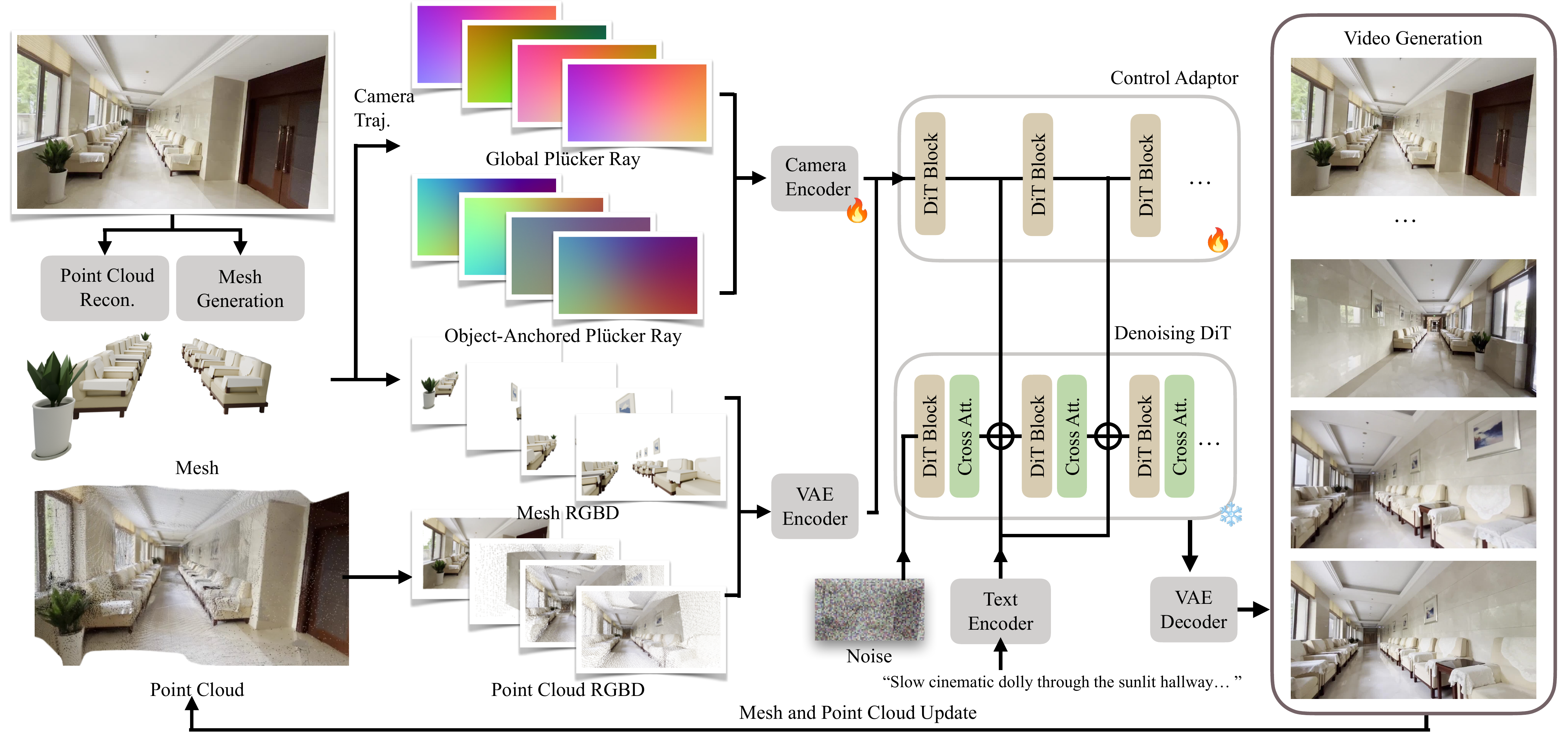}
    \captionsetup{font=normalsize}
    \vspace{-5px}
    \caption{\textbf{Pipeline Overview.} Given an image, \textbf{VivaWorld} renders extracted point clouds and posed meshes along the camera trajectory as visual guidance, and encodes camera motion with global and object-anchored Pl\"{u}cker rays to capture both scene-level motion and camera-object relationships. A compositional control adapter injects those signals into a pretrained video diffusion model, and generated frames update visual guidance to condition the next chunk for long-horizon generation.}
    \label{fig:pipeline}
    \vspace{-10px}
\end{figure*}

\section{Method}
Given a reference image $\mathbf{I}_{\mathrm{ref}}$, a text prompt, and a camera trajectory $\mathcal{C}=\{(K_t,E_t)\}_{t=1}^{T}$ with intrinsics $K_t$ and camera-to-world poses $E_t\in\mathrm{SE}(3)$, our goal is to generate a spatially and temporally consistent video $\mathcal{V}=\{\mathbf{I}_t\}_{t=1}^{T}$ that follows $\mathcal{C}$. As illustrated in Fig.~\ref{fig:pipeline}, we construct a compositional geometric representation $\mathcal{W}_{\mathrm{geo}}=\{\mathcal{P},\mathcal{M}\}$ of a background point cloud $\mathcal{P}$ and posed foreground meshes $\mathcal{M}$ (Sec.~\ref{sec:method:geometric-control}) and render it along $\mathcal{C}$ to obtain RGB-D conditions $\mathbf{c}_{\mathrm{rgbd}}$. We further introduce an object-anchored camera representation $\mathbf{c}_{\mathrm{cam}}$ encoding both global and object-relative camera motion (Sec.~\ref{sec:method:camera-control}). Both conditions are injected into a pretrained video diffusion transformer through lightweight adapters (Sec.~\ref{sec:method:model-design}). For long-horizon generation, we divide $\mathcal{C}$ into consecutive chunks $\mathcal{C}^{(q)}$ of up to $L$ poses and generate the corresponding video chunks $\mathcal{V}^{(q)}$ autoregressively, each conditioned on $\mathbf{c}_{\mathrm{rgbd}}^{(q)}$, $\mathbf{c}_{\mathrm{cam}}^{(q)}$, and previously generated frames as temporal context, while newly generated frames update $\mathcal{P}$ for subsequent chunks.

\vspace{-5px}
\subsection{Compositional Geometric Control}
\vspace{-5px}
\label{sec:method:geometric-control}
To preserve spatial consistency, we use a hybrid point-cloud and mesh representation $\mathcal{W}$ as geometric conditioning. We reconstruct a background point cloud $\mathcal{P}$ from observed regions and generate complete foreground meshes $\mathcal{M}$. Rendering $\mathcal{W}$ along the target camera trajectory $\mathcal{C}$ produces RGB-D conditions $\mathbf{c}_{\mathrm{rgbd}}$ for consistent video generation.

\textbf{Foreground Object Mesh Generation.}
To obtain object geometry that is both complete and aligned with the observations, we decompose posed-object generation into geometry generation and scene-space registration. We segment objects in $\mathbf{I}_{\mathrm{ref}}$ with SAM~3~\citep{carion2026sam} and generate a mesh $\widehat{\mathcal{M}}^k$ for each object $k$ using RecGen~\citep{zadaianchuk2026reconstruction} or GPT-6 Astra~\citep{openai2026gpt6astra}. We then align $\widehat{\mathcal{M}}^k$ to its observed points $\mathcal{P}^k_{\mathrm{obj}}$, extracted from the Pi3~\citep{wang2026pi} point cloud using the object's mask, via Iterative Closest Point (ICP) registration with per-axis scaling, estimating $T^k(\mathbf{x})=R^kD^k\mathbf{x}+\mathbf{t}^k$ with rotation $R^k\in\mathrm{SO}(3)$, translation $\mathbf{t}^k\in\mathbb{R}^3$, and anisotropic scale $D^k=\operatorname{diag}(s^k_x,s^k_y,s^k_z)$. The registered meshes form $\mathcal{M}=\{T^k(\widehat{\mathcal{M}}^k)\}_k$.

\begin{figure}[t]
\vspace{-5px}
\centering
\begin{minipage}[t]{0.45\textwidth}
\centering
\includegraphics[width=\linewidth]{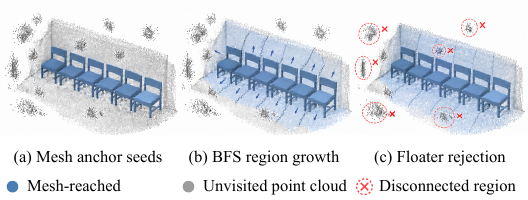}
\captionsetup{font=small,skip=2pt}
\caption{\textbf{Mesh-Seeded Point Cloud Filtering.} (a) Object meshes (blue) provide seeds. (b) BFS propagates through occupied voxels, leaving disconnected floating clusters unreached. (c) Unvisited components (red) are removed.}
\label{fig:mesh-seeded-ply-filter}
\end{minipage}\hfill
\begin{minipage}[t]{0.52\textwidth}
\centering
\includegraphics[width=\linewidth]{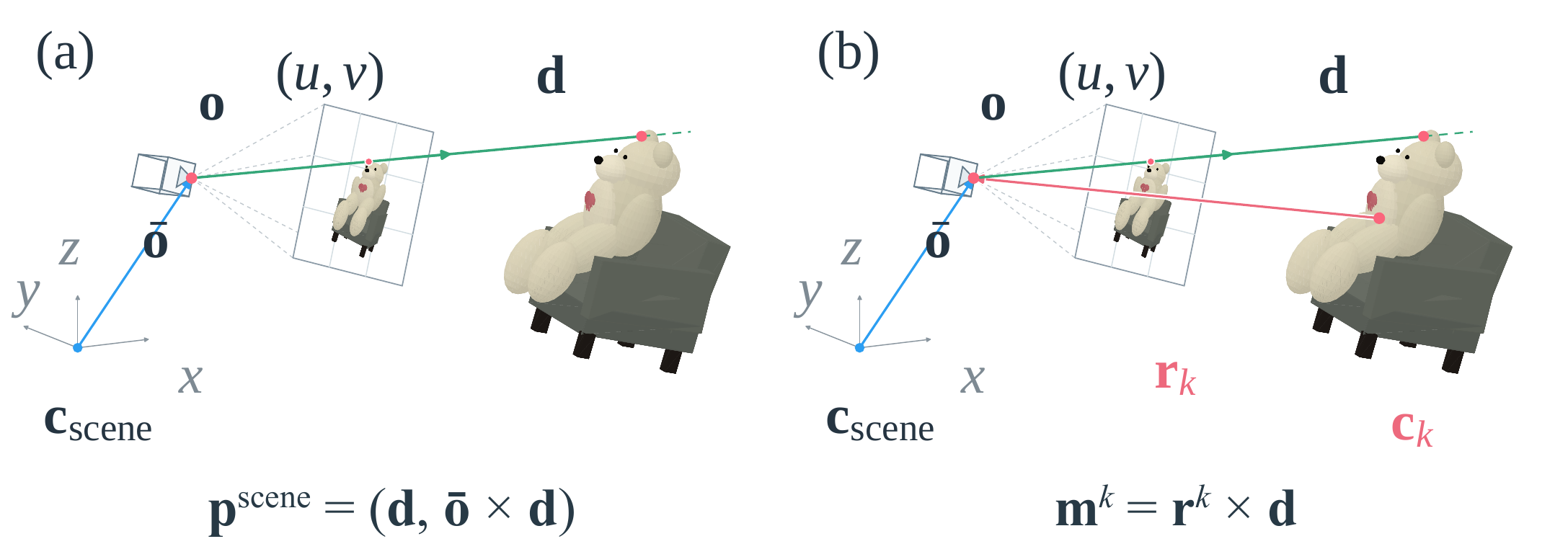}
\captionsetup{font=small,skip=2pt}
\caption{\textbf{Compositional Camera Representation.}
(a) The global ray $\mathbf{p}^{\mathrm{scene}}$ uses the camera offset from the scene center.
(b) Each object-anchored moment $\mathbf{m}^k$ is the cross product of the object-relative offset $\mathbf{r}^k$ from object center $\mathbf{c}^k$ and the ray direction $\mathbf{d}$.}
\label{fig:object-anchored-plucker}
\end{minipage}
\vspace{-15px}
\end{figure}

\textbf{Background Point Cloud Reconstruction.}
For each chunk, we use Pi3~\citep{wang2026pi} to predict dense metric depth and unproject it using the camera intrinsics and poses estimated by ViPE~\citep{huang2025vipe}, producing a local point cloud anchored at the chunk's first frame. Aggregating these local reconstructions yields a global point cloud $\mathcal{P}_{\mathrm{raw}}$. However, inconsistent depth estimates across chunks may introduce floating artifacts in mid-air that are disconnected from the main scene. To remove them, we propose a mesh-seeded connected-region filter (Fig.~\ref{fig:mesh-seeded-ply-filter}) using foreground meshes as geometric anchors. We voxelize $\mathcal{P}_{\mathrm{raw}}$ and $\mathcal{M}$, then flood-fill from mesh-occupied voxels through neighboring occupied voxels, and retain only points in reached voxels, removing all components disconnected from the registered meshes. The full algorithm is in Appendix~\ref{app:mesh-filter}.

\textbf{Compositional Visual Control.}
Given $\mathcal{W}_{\mathrm{geo}}=\{\mathcal{P},\mathcal{M}\}$, we render $\mathcal{P}$ and $\mathcal{M}$ separately along each chunk $\mathcal{C}^{(q)}$, yielding point-cloud and mesh RGB-D renderings $I^{\mathcal{P}}$ and $I^{\mathcal{M}}$ that form $\mathbf{c}_{\mathrm{rgbd}}^{(q)}$. We further derive a binary visual-guidance mask $\mathbf b$ from the union of the point-cloud and mesh rendering masks, indicating where either rendering provides visual guidance. The point cloud preserves observed appearance, while the meshes provide complete object geometry and clearer camera-motion cues through coherent view-dependent projections.
\vspace{-5px}
\subsection{Compositional Object-Oriented Camera Control}
\vspace{-5px}
\label{sec:method:camera-control}

Inspired by the observation that cameras typically move relative to objects and scenes, we propose a compositional camera embedding $\mathbf{c}_{\mathrm{cam}}$ combining the camera's global pose with its relative position to foreground objects to capture scene-level camera motion and object-centric viewpoint changes.

\textbf{Scene Camera Embedding.}
For frame $t$ with camera-to-world pose $E_t=[U_t\mid\mathbf{o}_t]$, where $U_t$ is the rotation and $\mathbf{o}_t$ the camera center, and intrinsics $K_t$, each pixel $(u,v)$ has world-space ray direction $\mathbf{d}_{t,u,v}=\operatorname{normalize}(U_tK_t^{-1}\widetilde{\mathbf{u}})$ with $\widetilde{\mathbf{u}}=(u,v,1)^\top$. Its Pl\"ucker coordinate is
\begin{equation} \mathbf{p}^{\mathrm{scene}}_{t,u,v}=\left(\mathbf{d}_{t,u,v},\bar{\mathbf{o}}_t\times\mathbf{d}_{t,u,v}\right),\qquad \bar{\mathbf{o}}_t=\frac{\mathbf{o}_t-\mathbf{c}_{\mathrm{scene}}}{s_{\mathrm{scene}}}, \label{eq:world-plucker} \end{equation}
where the scene center $\mathbf{c}_{\mathrm{scene}}$ and scale $s_{\mathrm{scene}}$ are computed from the camera and foreground-object centers. Stacking over all frames and pixels yields $\mathbf{p}^{\mathrm{scene}}\in\mathbb{R}^{T\times6\times H\times W}$.

\textbf{Object-Anchored Camera Embedding.}
Global Pl\"ucker rays do not explicitly capture camera motion relative to individual objects. We therefore use the registered meshes reconstructed from $\mathbf{I}_{\mathrm{ref}}$ as fixed object anchors with centers $\{\mathbf{c}^k\}_{k=1}^{N}$ in world coordinates. For each object $k$, we shift the ray origin to its center and compute the object-anchored moment as
\begin{equation} \mathbf{m}^{k}_{t,u,v}=\frac{\mathbf{o}_t-\mathbf{c}^k}{s_{\mathrm{scene}}}\times\mathbf{d}_{t,u,v},\qquad \mathbf{c}_{\mathrm{cam}}=\left[\mathbf{p}^{\mathrm{scene}}\middle\|\mathbf{m}^1\middle\|\cdots\middle\|\mathbf{m}^N\right]. \label{eq:object-anchored-camera} \end{equation}
Since $\mathbf{d}_{t,u,v}$ is shared across anchors, only the moment is added for each object, where $\|$ denotes channel-wise concatenation. The resulting tensor $\mathbf{c}_{\mathrm{cam}}\in\mathbb{R}^{T\times(6+3N)\times H\times W}$ jointly encodes global camera motion and object-relative viewpoint changes.
\vspace{-5px}
\subsection{Model Design}
\vspace{-5px}
\label{sec:method:model-design}
\textbf{Model Architecture.} We adopt the pretrained Wan~2.1 T2V-14B model as our frozen video diffusion backbone and build our compositional control adapter on VACE~\citep{jiang2025vace}. Given a target video chunk, the frozen Wan VAE encodes it into a latent $\mathbf{z}_0$, whose noisy version $\mathbf{z}_{\tau}$ at diffusion timestep $\tau$ is patchified into video tokens $\mathbf{x}_{\tau}$. The point-cloud and mesh RGB-D renderings $I^{\mathcal{P}}$ and $I^{\mathcal{M}}$ are encoded by the frozen Wan VAE, with RGB and depth encoded separately as $\mathcal{E}_{\mathrm{RGBD}}(I)=[\mathcal{E}_{\mathrm{VAE}}(I_{\mathrm{rgb}})\|\mathcal{E}_{\mathrm{VAE}}(I_{\mathrm{depth}})]$, while a lightweight spatiotemporal convolutional encoder $\Phi_{\mathrm{cam}}$ encodes the mask $\mathbf b$ and camera embedding $\mathbf{c}_{\mathrm{cam}}$. The resulting control latent is
\begin{equation} \mathbf{z}_{\mathrm{ctrl}}=\left[\mathcal{E}_{\mathrm{RGBD}}(I^{\mathcal{P}})\middle\|\mathcal{E}_{\mathrm{RGBD}}(I^{\mathcal{M}})\middle\|\Phi_{\mathrm{cam}}([\mathbf b\|\mathbf{c}_{\mathrm{cam}}])\right]. \label{eq:control-latent}
\end{equation}
The control latent is patchified into control tokens aligned one-to-one with $\mathbf{x}_{\tau}$ and processed by a transformer branch that injects residual features into selected blocks of the frozen Wan-DiT, sharing its timestep and text conditioning. We train only the adapter and camera encoder using the backbone's flow-matching objective. More architecture details are provided in Appendix~\ref{app:camera-encoder}.

\textbf{Supporting Single- and Multi-Image Inputs.}
For a single input image, we initialize the point cloud from $\mathbf{I}_{\mathrm{ref}}$ and update it causally using newly generated chunk-boundary frames. For sparse multi-image inputs, we use each image as the anchor frame of a corresponding video chunk and jointly reconstruct a global point cloud from all available views. During training, we simulate both settings (see Appendix~\ref{app:geo-cond} for training and data curation details).

\section{Experiments}
\vspace{-5px}
\subsection{Experimental Setup}
\vspace{-5px}
\paragraph{Implementation Details.} We build on VACE~\citep{jiang2025vace} with the frozen Wan~2.1-14B backbone~\citep{wan2025wan} and train only the compositional control adapter (3.05B parameters) and camera encoder on DL3DV-10K~\citep{ling2024dl3dv}. Data and more details are in Appendix.

\paragraph{Baselines and Evaluation Protocol.}
We evaluate our method from two perspectives: video generation and 3D scene reconstruction. For video generation, we compare against six representative camera-controlled methods: the static-scene video generation baselines AnyRecon~\citep{chen2026anyrecon}, Gen3C~\citep{ren2025gen3c}, Voyager~\citep{huang2025voyager}, and Lyra~2.0~\citep{shen2026lyra}, and the dynamic-scene video generation baselines VerseCrafter~\citep{zheng2026versecrafter} and NeoVerse~\citep{yang2026neoverse}. Among them, AnyRecon, Gen3C, and Lyra~2.0 support world-consistent long-video generation. We evaluate on 20 scenes ($\sim$ 200 video clips) from DL3DV-10K~\citep{ling2024dl3dv} test set and Tanks and Temples~\citep{knapitsch2017tanks} to assess in-domain performance and cross-dataset generalization. We evaluate both single-chunk and long-video generation at \(832\times480\) resolution. In the single-chunk setting, we explore ground-truth initial frames from chunks as the input and evaluate only the frames of each chunk common to all methods since chunk lengths differ. For long-video generation, we evaluate autoregressive predictions over the first 1,000 frames of each DL3DV scene and the first 240 frames of each Tanks and Temples scene. We report PSNR, SSIM, and LPIPS for visual fidelity and consistency and adapted WorldScore~\citep{duan2025worldscore} metrics for style and camera consistency. We also report rotation and unit-normalized translation errors (RotErr and TransErr) for camera-control accuracy. See more details in the appendix for the style metric and the camera metrics. For the quantitative study, we reconstruct 3D Gaussian splats from the generated videos using Depth Anything~3~\citep{lin2025depth} and evaluate the rendered views against the ground-truth images using SSIM and LPIPS. For the qualitative study,
we additionally explore 3DGS optimization for scene reoncstruction from generative videos.

\vspace{-5px}
\subsection{Quantitative Evaluation}
\vspace{-5px}
Tab.~\ref{tab:video_generation} compares single-chunk generation and long-video generation over 1,000 frames. Most baselines perform well on short clips but degrade substantially over longer horizons. NeoVerse achieves the highest short-clip style score, second best camera score and Voyager remains competitive on short clips, but both suffer large declines in visual quality over long horizons, with Voyager degrading a lot. VerseCrafter and AnyRecon exhibit the weakest long-video camera control, while Gen3C maintains strong style consistency but accumulates camera error. Lyra2 is the strongest baseline overall. Our method achieves the best PSNR, SSIM, and LPIPS in both settings and the best long-video camera-control score, while Gen3C achieves the highest long-video style score. As shown in Tab.~\ref{tab:camera_scene}, our method reduces long-video rotation and translation errors by \(23.4\%\) and \(38.8\%\), respectively, compared with Lyra2, demonstrating that our hybrid 3D guidance and object-anchored camera embedding effectively mitigate accumulated drift. For downstream 3D scene generation, our method achieves the best SSIM and LPIPS in both settings, whereas all baselines show reduced reconstruction quality as video length increases. By incorporating object anchors into both visual guidance and camera embeddings, our method preserves stronger cross-frame geometric consistency, resulting in more accurate and coherent 3D reconstructions.

\begin{table*}[t]
\centering
\vspace{-5px}
\caption{\textbf{Video Generation Scores.} Baselines degrade sharply from single
chunks to long videos, while VivaWorld achieves the best visual fidelity in both and
the best long-video camera control.}
\vspace{-5px}
\label{tab:video_generation}
\setlength{\tabcolsep}{3pt}
\renewcommand{\arraystretch}{1.15}

\small
\resizebox{\textwidth}{!}{\begin{tabular}{lccccc|ccccc}
\hline
&
\multicolumn{5}{c|}{\textbf{Single-Chunk Generation}}
&
\multicolumn{5}{c}{\textbf{Long-Video Generation}}
\\[4pt]
\cline{2-11}
\textbf{Methods}
&
PSNR$\uparrow$
&
SSIM$\uparrow$
&
LPIPS$\downarrow$
&
Style$\uparrow$
&
Cam.$\uparrow$
&
PSNR$\uparrow$
&
SSIM$\uparrow$
&
LPIPS$\downarrow$
&
Style$\uparrow$
&
Cam.$\uparrow$
\\[2pt]
\hline

AnyRecon
& 18.54 & 0.436 & 0.285 & 98.67 & 94.75
& 12.64 & \cellcolor{ScoreSecond}{0.361} & 0.600 & 96.37 & 27.37 \\

Gen3C
& 19.92 & 0.513 & 0.254 & 98.09 & 91.61
& 12.71 & 0.333 & 0.609 & \cellcolor{Palette4}{\textbf{96.99}} & 62.56 \\

Voyager
& 18.60 & 0.446 & 0.345 & 98.62 & 92.57
& 8.622 & 0.213 & 0.796 & 91.96 & 55.58 \\

Lyra2
& \cellcolor{ScoreSecond}{20.01} & \cellcolor{ScoreSecond}{0.528} & \cellcolor{ScoreSecond}{0.227} & \cellcolor{ScoreSecond}{98.74} & \cellcolor{Palette4}{\textbf{96.34}}
& \cellcolor{ScoreSecond}{12.85} & 0.343 & \cellcolor{ScoreSecond}{0.582} & \cellcolor{ScoreSecond}{96.82} & \cellcolor{ScoreSecond}{65.09} \\

VerseCrafter
& 17.24 & 0.431 & 0.292 & 97.45 & 85.53
& 11.00 & 0.294 & 0.691 & 95.48 & 27.76 \\

NeoVerse
& 17.70 & 0.415 & 0.370 & \cellcolor{Palette4}{\textbf{99.02}} & \cellcolor{ScoreSecond}{96.22}
& 10.42 & 0.243 & 0.704 & 93.67 & 49.09 \\

\hline

Ours
& \cellcolor{Palette4}{\textbf{21.24}} & \cellcolor{Palette4}{\textbf{0.570}} & \cellcolor{Palette4}{\textbf{0.173}} & 98.65 & 95.22
& \cellcolor{Palette4}{\textbf{13.92}} & \cellcolor{Palette4}{\textbf{0.376}} & \cellcolor{Palette4}{\textbf{0.529}} & 96.14 & \cellcolor{Palette4}{\textbf{79.51}} \\

\hline
\vspace{-5px}
\end{tabular}
}

\renewcommand{\arraystretch}{1.0}
\vspace{-5pt}
\end{table*}

\begin{table*}[t]
\centering
\vspace{-5px}
\caption{\textbf{Camera Control and 3D Scene Generation.}
VivaWorld achieves the lowest long-video camera errors and the most accurate 3D reconstructions.}
\vspace{-5px}
\setlength{\tabcolsep}{1.8pt}
\renewcommand{\arraystretch}{1.15}
\label{tab:camera_scene}
\small
\resizebox{\textwidth}{!}{\begin{tabular}{@{}lcc|cc||cc|cc@{}}
\hline

&
\multicolumn{4}{c||}{\textbf{Camera Control}}
&
\multicolumn{4}{c}{\textbf{3D Scene Generation}}
\\[2pt]

\cline{2-9}

&
\multicolumn{2}{c|}{\textbf{Single-Chunk Gen.}}
&
\multicolumn{2}{c||}{\textbf{Long-Video Gen.}}
&
\multicolumn{2}{c|}{\textbf{Single-Chunk Gen.}}
&
\multicolumn{2}{c}{\textbf{Long-Video Gen.}}
\\[2pt]

\cline{2-9}

&
RotErr$\downarrow$
&
TransErr$\downarrow$
&
RotErr$\downarrow$
&
TransErr$\downarrow$
&
SSIM$\uparrow$
&
LPIPS$\downarrow$
&
SSIM$\uparrow$
&
LPIPS$\downarrow$
\\

\hline

AnyRecon
& 0.375 & 0.036
& 28.47 & 0.212 & 0.461 & 0.449 & \cellcolor{ScoreSecond}{0.444} & 0.673 \\

Gen3C
& 0.655 & 0.056
& 14.30 & 0.121 & 0.472 & 0.448 & 0.418 & 0.676 \\

Voyager
& 0.452 & 0.051
& 14.91 & 0.191 & 0.461 & 0.492 & 0.350 & 0.802 \\

Lyra2
& \cellcolor{Palette4}{\textbf{0.215}} & \cellcolor{ScoreSecond}{0.025}
& \cellcolor{ScoreSecond}{9.022} & \cellcolor{ScoreSecond}{0.098} & \cellcolor{ScoreSecond}{0.472} & \cellcolor{ScoreSecond}{0.439} & 0.423 & \cellcolor{ScoreSecond}{0.658} \\

VerseCrafter
& 0.604 & 0.107
& 28.91 & 0.257 & 0.451 & 0.475 & 0.403 & 0.763 \\

NeoVerse
& 0.400 & \cellcolor{Palette4}{\textbf{0.023}}
& 22.14 & 0.189 & 0.470 & 0.460 & 0.415 & 0.781 \\

\hline

Ours
& \cellcolor{ScoreSecond}{0.308} & 0.033
& \cellcolor{Palette4}{\textbf{6.908}} & \cellcolor{Palette4}{\textbf{0.060}} & \cellcolor{Palette4}{\textbf{0.480}} & \cellcolor{Palette4}{\textbf{0.417}} & \cellcolor{Palette4}{\textbf{0.447}} & \cellcolor{Palette4}{\textbf{0.598}} \\

\hline
\end{tabular}
}

\renewcommand{\arraystretch}{1.0}
\vspace{-15pt}
\end{table*}

\vspace{-5px}
\subsection{Qualitative Evaluation}
\begin{figure*}[!t]
\centering

\includegraphics[width=\textwidth,trim=0 0 0 749,clip]{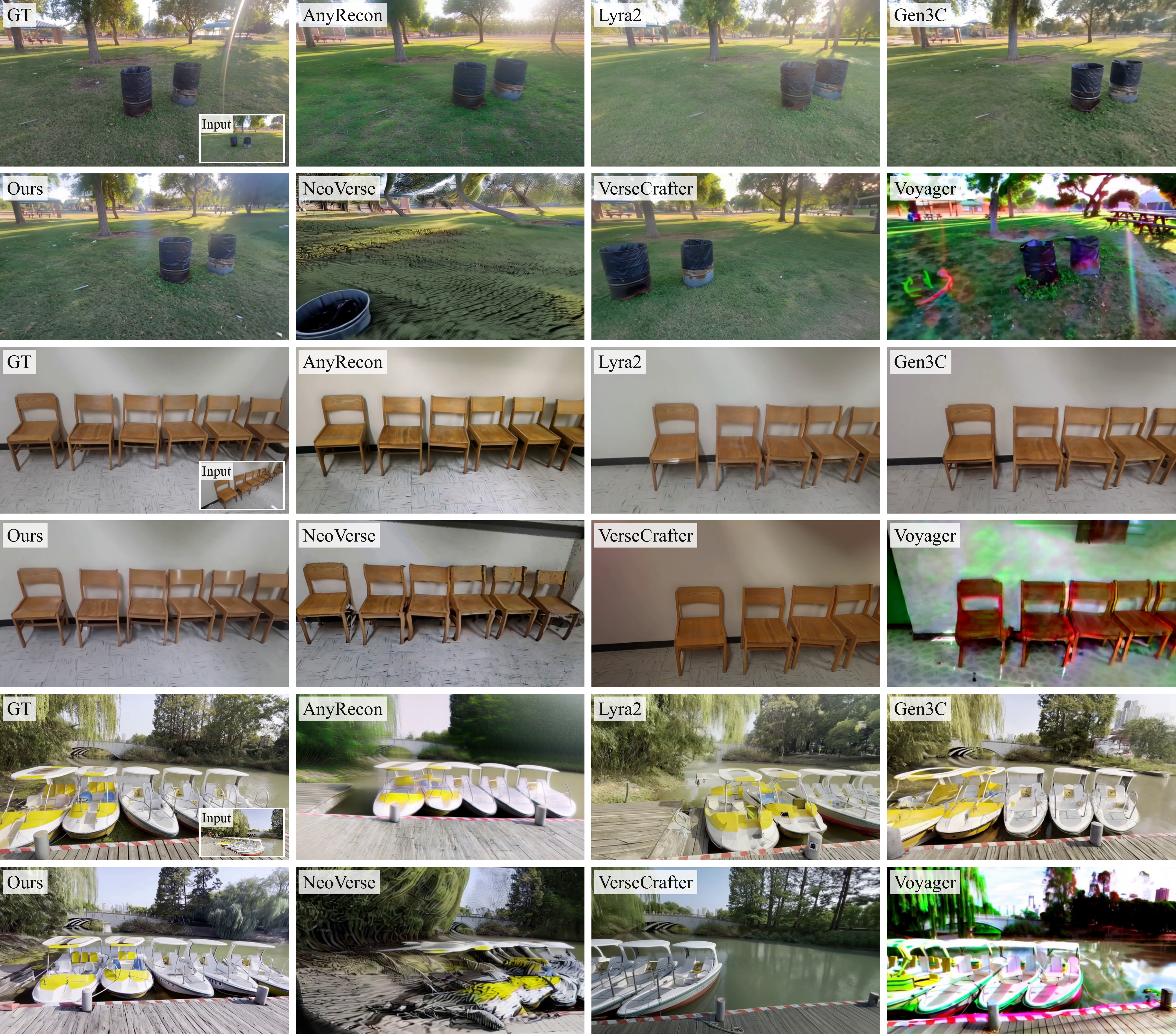}
\vspace{-15px}
\caption{\textbf{Video Generation.} Baselines show color artifacts, distortion, or incorrect object counts, while \textbf{VivaWorld} preserves the ground-truth layout and appearance.}
\label{fig:qualitative_video_comparison}
\par\vspace{6pt}
\centering
\includegraphics[width=\textwidth]{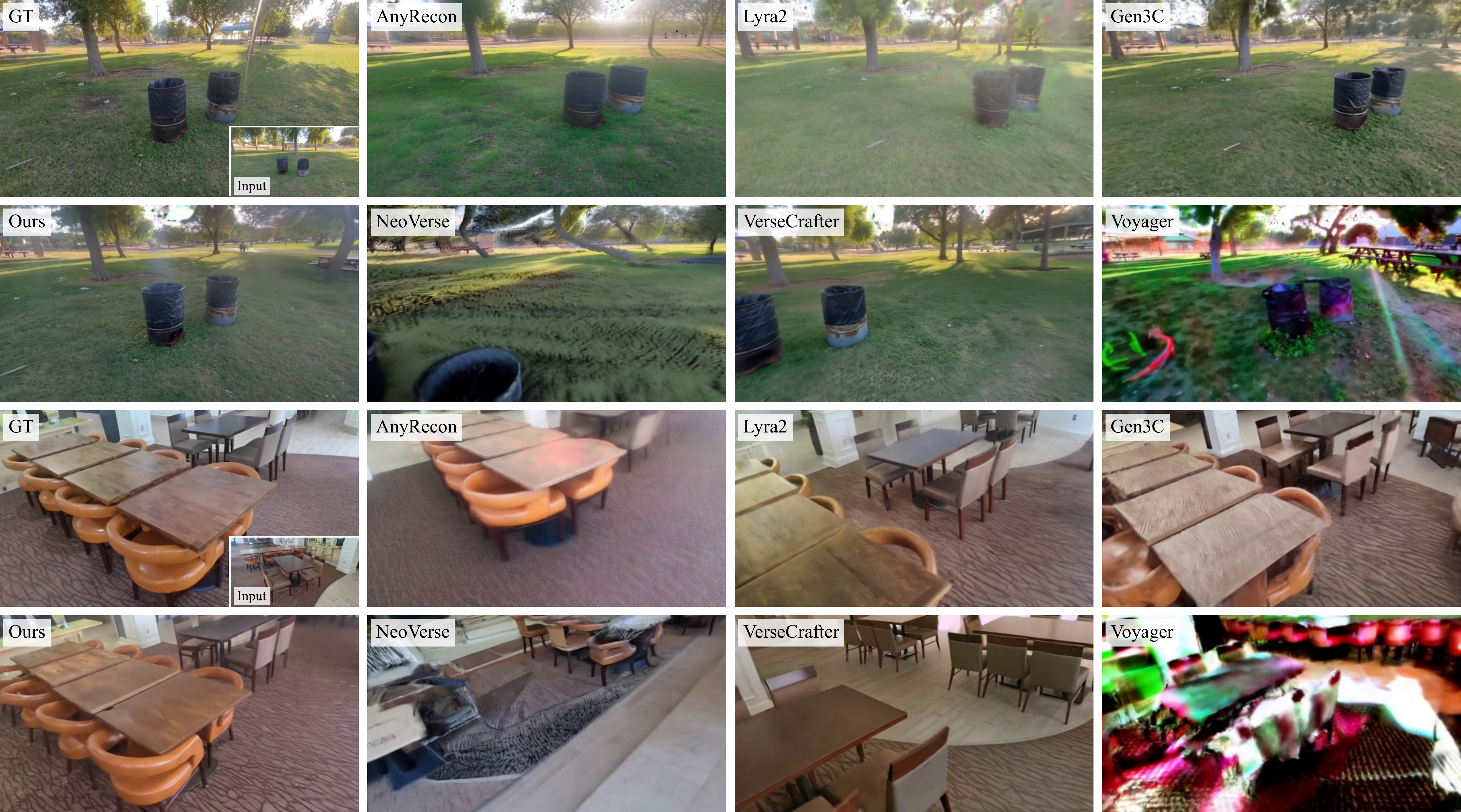}
\caption{\textbf{3D Reconstruction.} We reconstruct 3D scenes from generated videos with per-scene 3DGS optimization. Baselines produce blurry geometry, while ours yields a clean reconstruction.}
\vspace{-16pt}
\label{fig:qualitative_scene_comparison}
\end{figure*}

\vspace{-5px}
Fig.~\ref{fig:qualitative_video_comparison} presents qualitative comparisons among video generation methods. NeoVerse, VerseCrafter, and Voyager perform well over short horizons but degrade over longer ones: NeoVerse propagates reconstruction artifacts under large camera motion, while VerseCrafter and Voyager lose global spatial consistency, producing incorrect object counts (e.g., yachts or chairs). Lyra 2.0, Gen3C, and AnyRecon maintain better long-horizon quality but still degrade: Lyra 2.0 and Gen3C accumulate camera drift and lose texture detail, and AnyRecon becomes increasingly blurry. In contrast, our method accurately follows the target trajectory while preserving appearance and geometric consistency throughout. For downstream 3D reconstruction from generated videos (Fig.~\ref{fig:qualitative_scene_comparison}), we evaluate using optimization-based reconstruction for 3000 steps. For baseline methods, inconsistent object positions across generated frames lead to duplicated or blurry geometry, and objects can even take on a different shape from the input frame. In contrast, our method produces more accurate and coherent reconstructions. Additional results are provided in the appendix and supplementary video.

\clearpage

\begin{minipage}{\textwidth}
\centering
\captionsetup{skip=2pt}
\includegraphics[width=\textwidth]{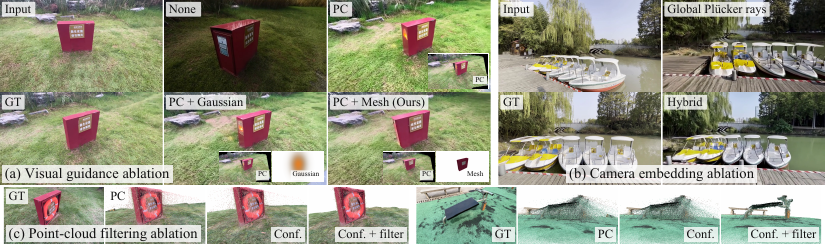}
\captionof{figure}{\textbf{Ablation.} Ablation studies showing our full configuration outperforms alternative settings. }

\label{fig:ablation-studies}
\label{fig:visual_guidance_ablation}
\label{fig:camera-embedding-ablation}
\label{fig:point-filter-ablation}
\end{minipage}

\vspace{4pt}
\begin{minipage}{\textwidth}
\centering
\captionof{table}{\textbf{Ablation Study.} The ablation study indicates that exploring hybrid visual guidance and composition yields consistent gains, achieving better performance than the other settings.}
\label{tab:ablation}
\label{tab:ablation_visual_guidance}
\label{tab:guidance_ablation}
\small
\setlength{\tabcolsep}{5pt}
\begin{tabular}{lcccccc}
\toprule
& \multicolumn{3}{c}{Single-Chunk}
& \multicolumn{3}{c}{Long-Video} \\
\cmidrule(lr){2-4} \cmidrule(lr){5-7}
Guidance Signal
& PSNR $\uparrow$ & LPIPS $\downarrow$ & Cam $\uparrow$
& PSNR $\uparrow$ & LPIPS $\downarrow$ & Cam $\uparrow$ \\
\midrule
None
& 11.47 & 0.490 & 60.35 & 10.03 & 0.700 & 3.17 \\
PC
& 20.00 & 0.174 & 89.66 & 12.41 & 0.584 & 71.34 \\
PC + Gaussian
& \cellcolor{ScoreSecond}{20.76} & 0.168 & 83.94 & \cellcolor{ScoreSecond}{13.60} & 0.550 & \cellcolor{ScoreSecond}{79.95} \\
PC + Mesh + Global Embedding
& 20.49 & \cellcolor{ScoreSecond}{0.149} & \cellcolor{ScoreSecond}{89.84} & 13.56 & \cellcolor{ScoreSecond}{0.512} & 78.32 \\
\textbf{Ours (PC + Mesh + Hybrid Embedding)}
& \cellcolor{Palette4}{\textbf{21.62}} & \cellcolor{Palette4}{\textbf{0.145}} & \cellcolor{Palette4}{\textbf{90.69}}
& \cellcolor{Palette4}{\textbf{14.30}} & \cellcolor{Palette4}{\textbf{0.488}} & \cellcolor{Palette4}{\textbf{80.97}} \\
\bottomrule
\end{tabular}
\end{minipage}

\subsection{Ablation}
We ablate the key components of the proposed framework in the DL3DV test set.

\textbf{3D Compositional Guidance.}
We compare the hybrid mesh and point-cloud guidance against no visual guidance (``None''), point clouds alone (``PC''), and point clouds augmented with object-level 3D Gaussian proxies (``PC+Gaussian'') as in~\citep{zheng2026versecrafter} (Tab.~\ref{tab:ablation_visual_guidance}). All guided variants outperform the unguided one, which suffers substantial camera drift and deviation from the target trajectory in long-video generation. Point clouds improve camera alignment by providing global 3D memory, and Gaussian proxies further aid camera control by making camera motion easier to perceive, particularly over repetitive background textures such as grass. Integrated point cloud guidance with meshes improves not only camera consistency but also image quality via stronger geometry priors.

\textbf{Camera Embedding.} Tab.~\ref{tab:ablation} compares different Plücker-ray embeddings. In Fig.~\ref{fig:camera-embedding-ablation}, hybrid embedding improves visual quality and controllability.

\textbf{Filtered Point Cloud.} In Fig.~\ref{fig:ablation-studies}(c), we compare the reconstructed point cloud without confidence score and connection filtering, the point cloud after confidence score filtering, and our point cloud guidance using both confidence scores and the proposed connectivity-based filtering. Confidence-based filtering removes some artifacts but leaves substantial floating points, particularly near object boundaries. Our method removes floating points near foreground objects while suppressing background noise, providing cleaner visual guidance.

\clearpage

\begin{figure*}[!t]
\captionsetup{skip=2pt}
\centering
\includegraphics[width=\textwidth]{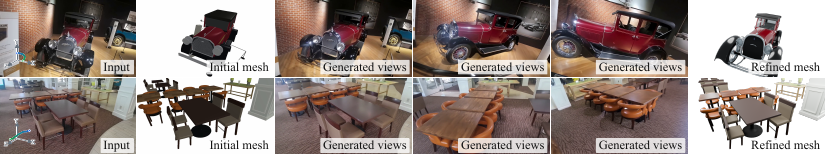}
\caption{\textbf{Agentic Mesh and Video Generation.} Given an image and an initial mesh, our method generates novel views that provide visual references for refining agentic generation.}
\label{fig:agentic_generation}

\par\vspace{6pt}
\centering
\includegraphics[width=\textwidth]{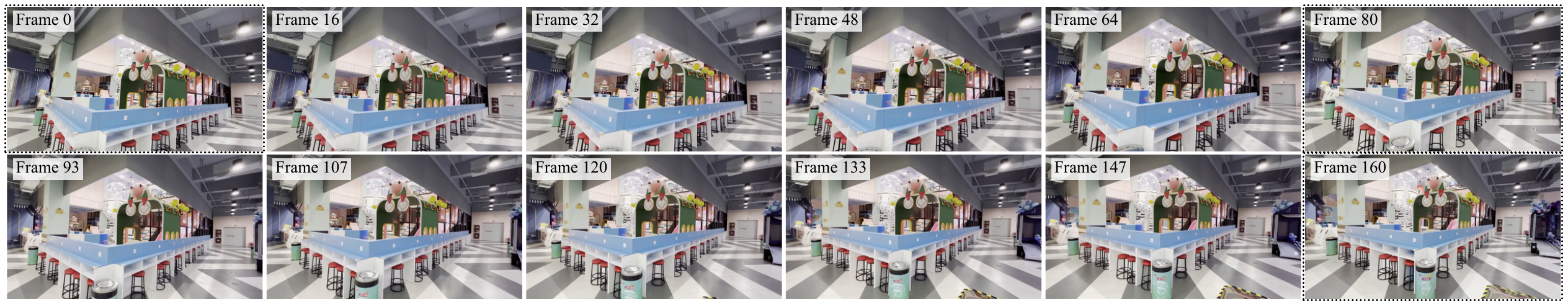}
\caption{\textbf{Video Interpolation.} Our method generates intermediate frames from keyframes.}
\label{fig:video_interpolation}
\end{figure*}

\subsection{Applications}
\textbf{Agentic Video Generation.}
We support agentic 3D objects as compositional guidance, enabling iterative 3D-video co-refinement. In Fig.~\ref{fig:agentic_generation}, a GPT-generated mesh guides video generation along the target camera trajectory, and the resulting video refines the mesh quality and spatial arrangement.

\textbf{Video Interpolation.}
We also support chunk-wise high-quality video interpolation between sparse input views, using a fixed global point cloud and object meshes as geometric guidance (Fig.~\ref{fig:video_interpolation}).

\textbf{Humanoid Robot Learning.}
The generated videos enable compositional 3D reconstruction, supporting humanoid robot learning of walking and sitting tasks (Fig.~\ref{fig:teaser}).

\section{Conclusion, Limitation and Future Work}
We presented \textbf{VivaWorld}, a novel framework for generating long, 3D-grounded videos with spatial consistency and accurate camera control from a single image. By integrating point-cloud reconstruction priors with generative mesh priors as hybrid visual guidance and introducing compositional camera conditioning, VivaWorld produces spatially coherent video frames over extended trajectories. However, the current pipeline does not fully prevent color drift during long-horizon generation. Future work could explore explicit appearance-consistency constraints or color alignment strategies to preserve consistent colors throughout the long generation video.

\bibliography{references}
\bibliographystyle{plainnat}

\appendix

\newpage
\setlength{\parskip}{3pt}
\setlength{\floatsep}{6pt plus 2pt minus 2pt}
\setlength{\textfloatsep}{8pt plus 2pt minus 2pt}
\setlength{\intextsep}{8pt plus 2pt minus 2pt}
\captionsetup{skip=5pt}
\section{Method \& Training Details}
\label{app:method}

\subsection{Mesh-Seeded Point Cloud Filtering}
\label{app:mesh-filter}
Given the raw point cloud $\mathcal{P}_{\mathrm{raw}}$ and foreground meshes $\mathcal{M}$, we voxelize both into a grid of resolution $R\times R\times R$. Let $\mathcal{O}_R=\operatorname{Vox}_R(\mathcal{P}_{\mathrm{raw}})\cup\operatorname{Vox}_R(\mathcal{M})$ denote all occupied voxels and $\mathcal{S}_R=\operatorname{Vox}_R(\mathcal{M})$ the mesh-occupied seed voxels. Starting from $\mathcal{S}_R$, we perform multi-source flood filling over $\mathcal{O}_R$. At each step, an occupied voxel $\mathbf{v}$ expands to neighboring occupied voxels $\mathbf{u}$ satisfying $0<\|\mathbf{u}-\mathbf{v}\|_\infty\leq r$. Denoting all visited voxels by $\mathcal{R}_r$, we obtain the filtered point cloud as
\begin{equation}
\mathcal{P}
=
\left\{
\mathbf{p}\in\mathcal{P}_{\mathrm{raw}}
\;\middle|\;
\operatorname{Vox}_R(\mathbf{p})\in\mathcal{R}_r
\right\},
\label{eq:mesh-seeded-filter}
\end{equation}
where $\operatorname{Vox}_R(\mathbf{p})$ returns the voxel containing $\mathbf{p}$. This operation retains scene geometry connected to at least one registered mesh and removes all unvisited components as floaters.

We evaluate conditioning quality across five cases by comparing renderings of progressively reconstructed point clouds with ground-truth images. Table~\ref{tab:mesh_seeded_filtering} reports mean PSNR gains over unfiltered point clouds. Confidence filtering at a threshold of $0.01$ improves conditioning quality by suppressing reconstruction noise, while mesh-seeded filtering achieves the largest gain with a $128 \times 128 \times 128$ voxel grid and $r=1$. We nevertheless choose $r=2$, which achieves a nearly identical gain while allowing connections across a one-voxel gap, providing greater tolerance to fragmented geometry.

\begin{table*}[htbp]
\centering
\caption{\textbf{Mesh-seeded Filtering Ablation.} Mean conditioning PSNR gains (dB) over unfiltered point clouds across five cases. Grid size denotes the voxel resolution along each axis, and $r$ denotes the connectivity radius.}
\label{tab:mesh_seeded_filtering}
\small
\setlength{\tabcolsep}{8pt}
\begin{tabular}{ccccccc}
\toprule
Confidence
& \multicolumn{2}{c}{Grid $32^3$}
& \multicolumn{2}{c}{Grid $64^3$}
& \multicolumn{2}{c}{Grid $128^3$} \\
\cmidrule(lr){2-3}
\cmidrule(lr){4-5}
\cmidrule(lr){6-7}
$\geq 0.01$
& $r=1$ & $r=2$
& $r=1$ & $r=2$
& $r=1$ & $r=2$ (Ours) \\
\midrule
$+0.0512$
& $+0.0028$ & $+0.0080$
& $+0.0532$ & $+0.0531$
& $\mathbf{+0.1402}$ & $\mathbf{+0.1387}$ \\
\bottomrule
\end{tabular}
\end{table*}

In addition to using a connectivity radius of $r=2$, we use multiple object meshes derived from SAM's ``everything'' mode as seeds to preserve valid regions disconnected by occlusion or sparse sampling, increasing coverage across the scene. We acknowledge that mesh-seeded filtering parameters, including voxel size and connectivity radius, may require adjustment for complex scenes or inaccurate reconstructions. Each filtering run takes approximately one second on average, making it practical to evaluate several parameter settings. Selecting the best-performing setting for each case yields a mean PSNR gain of $+0.2421$~dB across these five cases.

\subsection{Camera Encoder}
\label{app:camera-encoder}
The camera encoder $\Phi_{\mathrm{cam}}$ is a lightweight single-layer 3D CNN that takes the concatenation of the visual-guidance mask $\mathbf{b}$ and the camera embedding $\mathbf{c}_{\mathrm{cam}}$ (16 channels for $N=3$ anchors) and applies one $\operatorname{Conv3D}$ with kernel size and stride $(4,8,8)$, followed by GroupNorm and SiLU, producing a 64-channel feature aligned with the $4\times$ temporal and $8\times$ spatial compression of the Wan VAE latent. Since the camera embedding is already a dense, pixel-aligned geometric signal, this single layer is sufficient, and the encoder contains only 0.26M parameters. By default, we consider three object candidates as anchors. If mesh generation produces fewer than three meshes, we pad the set to three and if it produces more than three, we select the three 3D objects most similar to the segmented image, based on image–object similarity.

\subsection{Geometric Conditioning}
\label{app:geo-cond}
To support both single- and multi-image inputs, we curate two complementary point-cloud conditions. For multi-image conditioning, we uniformly sample 20 frames across the video, use each as the anchor of a corresponding chunk, and jointly reconstruct a global point cloud $\mathcal{P}^{\mathrm{global}}$ shared across all chunks. For single-image autoregressive generation, we construct a causal point cloud $\mathcal{P}^{(q)}_{\mathrm{causal}}$ for each chunk $q$ using only $\mathbf{I}_{\mathrm{ref}}$ and the anchor frames of preceding chunks, so the first chunk uses only $\mathbf{I}_{\mathrm{ref}}$ and later chunks progressively incorporate available observations without accessing future frames. During training, we sample global and causal conditions with equal probability, exposing the model to both complete multi-view geometry and the partial geometry available during autoregressive inference. We further augment conditions by using both raw and filtered point clouds, with a voxel resolution of $R=128$ and a connectivity radius of $r=2$ for floater removal. Foreground meshes are generated with RecGen~\citep{zadaianchuk2026reconstruction}, and 20\% of training samples are augmented with GPT-6 Astra~\citep{openai2026gpt6astra} agentic meshes.

\subsection{Data Curation}
\label{app:data-curation}
We train on DL3DV-10K~\citep{ling2024dl3dv}, retaining every third frame, which yields approximately 1,000-1,500 frames per video. Videos are divided into 81-frame chunks, with adjacent chunks sharing one boundary frame. Camera intrinsics and poses are estimated with ViPE~\citep{huang2025vipe}, and all reconstructed geometry is aligned within the resulting shared scene coordinate system. Point clouds are reconstructed with Pi3~\citep{wang2026pi}, and video captions are generated with GPT-6. The point clouds and meshes are rendered along the estimated camera trajectories to produce RGB-D conditions paired with the target frames. We processed full 8,965 scenes from DL3DV using RecGen and a subset of approximately 5,246 scenes using Astra. Astra failed to produce any object meshes for only 0.4\% of the processed scenes, while the corresponding failure rate for RecGen was approximately 3.5\%. Using both mesh-generation methods reduced the proportion of scenes without any meshes to less than 0.3\%, which means that over 99.7\% of scenes have at least one mesh available for conditioning.
\begin{wraptable}[19]{r}{0.4\textwidth}
    \centering
    \small
    \setlength{\tabcolsep}{4pt}
    \renewcommand{\arraystretch}{0.90}
    \caption{\textbf{Mesh Curation Analysis.}}
    \label{tab:mesh_counts}
    \begin{tabular}{@{}lrr@{}}
        \toprule
        Count per scene & RecGen & Astra \\
        \midrule
        0        & 320   & 23 failures \\
        1        & 407   & 0     \\
        2        & 443   & 4     \\
        3        & 510   & 6     \\
        4        & 458   & 5     \\
        5        & 521   & 18    \\
        6        & 491   & 25    \\
        7        & 466   & 28    \\
        8        & 413   & 48    \\
        9        & 402   & 61    \\
        10       & 391   & 87    \\
        11--15   & 1,673 & 802   \\
        16--20   & 1,187 & 1,233 \\
        21--30   & 1,007 & 1,800 \\
        31--50   & 252   & 811   \\
        51--100  & 24    & 214   \\
        101--200 & 0     & 81    \\
        201--429 & 0     & 23    \\
        \midrule
        Counted scenes & 8,965 & 5,246 \\
        Mean count     & 11.39 & 26.94 \\
        Maximum count  & 83    & 429   \\
        \bottomrule
    \end{tabular}
    \vspace{-\intextsep}
\end{wraptable}

\subsection{Training Hyperparameters}
\label{app:training}
We finetune the control adapter with the flow-matching velocity loss, using 1,000 training timesteps with a schedule shift of 5.0. We use AdamW with a learning rate of $1\times10^{-5}$, a linear warmup over the first 100 iterations followed by a constant learning rate, $(\beta_1,\beta_2)=(0.9,0.999)$, a weight decay of 0.01, and gradient clipping at a norm of 1.0. To support classifier-free guidance, we drop the text prompt with probability 0.1. We also drop the visual controls with probability 0.05. Training runs on 64 NVIDIA A100 GPUs, for a global batch size of 64 for approximately 10000 iterations.

\begin{table*}[!b]
\centering
\caption{\textbf{Robustness Analysis.}}
\label{tab:upstream_sensitivity}
\normalsize
\setlength{\tabcolsep}{4pt}
\begin{tabular*}{\textwidth}{@{\extracolsep{\fill}}lcccc@{}}
\toprule
Configuration
& Conditioning PSNR $\uparrow$
& PSNR $\uparrow$
& SSIM $\uparrow$
& LPIPS $\downarrow$ \\
\midrule
1 anchor
& 11.66 & 14.26 & 0.3617 & 0.5161 \\
2 anchors
& 11.74 & 14.36 & 0.3625 & \textbf{0.5122} \\
\textbf{Full Mesh + Full PLY (3 anchors)}
& \textbf{11.81} & \textbf{14.41}
& \textbf{0.3658} & 0.5141 \\
Half Mesh + Full PLY
& 11.69 & 14.33 & 0.3616 & 0.5231 \\
Full Mesh + PLY (conf.\ $\geq 0.3$)
& 9.490 & 14.13 & 0.3540 & 0.5229 \\
Mesh Position Shift of 5\% + Full PLY
& 11.71 & 14.18 & 0.3564 & 0.5310 \\
\bottomrule
\end{tabular*}
\end{table*}

\begin{table*}[!t]
\centering
\caption{\textbf{Runtime and Memory Comparison.}}
\label{tab:runtime_comparison}
\normalsize
\setlength{\tabcolsep}{6pt}
\begin{tabular*}{\textwidth}{@{\extracolsep{\fill}}lccc@{}}
\toprule
Method
& Model Time (min) $\downarrow$
& Total Time (min) $\downarrow$
& Peak Memory (GiB) $\downarrow$ \\
\midrule
Ours         & 157.6 & 166.4 & 44.76 \\
Gen3C        & \textbf{38.9} & \textbf{41.0} & \textbf{21.98} \\
AnyRecon     & 50.6  & 54.1  & 40.38 \\
VerseCrafter & 148.1 & 172.6 & 54.87 \\
Lyra2        & 208.7 & 225.9 & 66.73 \\
NeoVerse     & 239.1 & 254.8 & 46.54 \\
Voyager      & 457.1 & 514.3 & 42.28 \\
\bottomrule
\end{tabular*}
\end{table*}

\begin{table*}[!t]
\centering
\caption{\textbf{Runtime breakdown.}}
\label{tab:runtime_breakdown}
\normalsize
\setlength{\tabcolsep}{5pt}
\renewcommand{\arraystretch}{1.1}
\begin{tabular*}{\textwidth}{@{\extracolsep{\fill}}>{\raggedright\arraybackslash}p{0.30\textwidth}r>{\raggedright\arraybackslash}p{0.50\textwidth}@{}}
\toprule
Component / Step & Time (min) & Description / Details \\
\midrule
Video diffusion denoising
& 150.3
& 13 chunks, 30 denoising steps each \\
Other processing across chunks
& 7.31
& Input preparation, text/VAE processing, video encoding and saving \\
Processing outside chunks
& 4.06
& Model initialization, initial and cumulative Pi3 control updates, stitching and validation \\
First-frame Pi3
& 1.79
& Including model loading \\
First-frame SAM3 segmentation
& 0.74
& Including model loading \\
First-frame RecGen
& 1.93
& Model loading and mesh generation \\
Voxel-based filtering
& 0.30
& Point-cloud filtering \\
\midrule
\textbf{Total}
& \textbf{166.4}
& \textbf{Approximately 2 hours 46 minutes} \\
\midrule
Astra: 1 iteration (optional)
& 2.89
& Excluded from the total \\
Astra: 2 iterations (optional)
& 6.23
& Excluded from the total \\
\bottomrule
\end{tabular*}
\end{table*}

\begin{figure*}[!t]
\centering
\includegraphics[width=\textwidth]{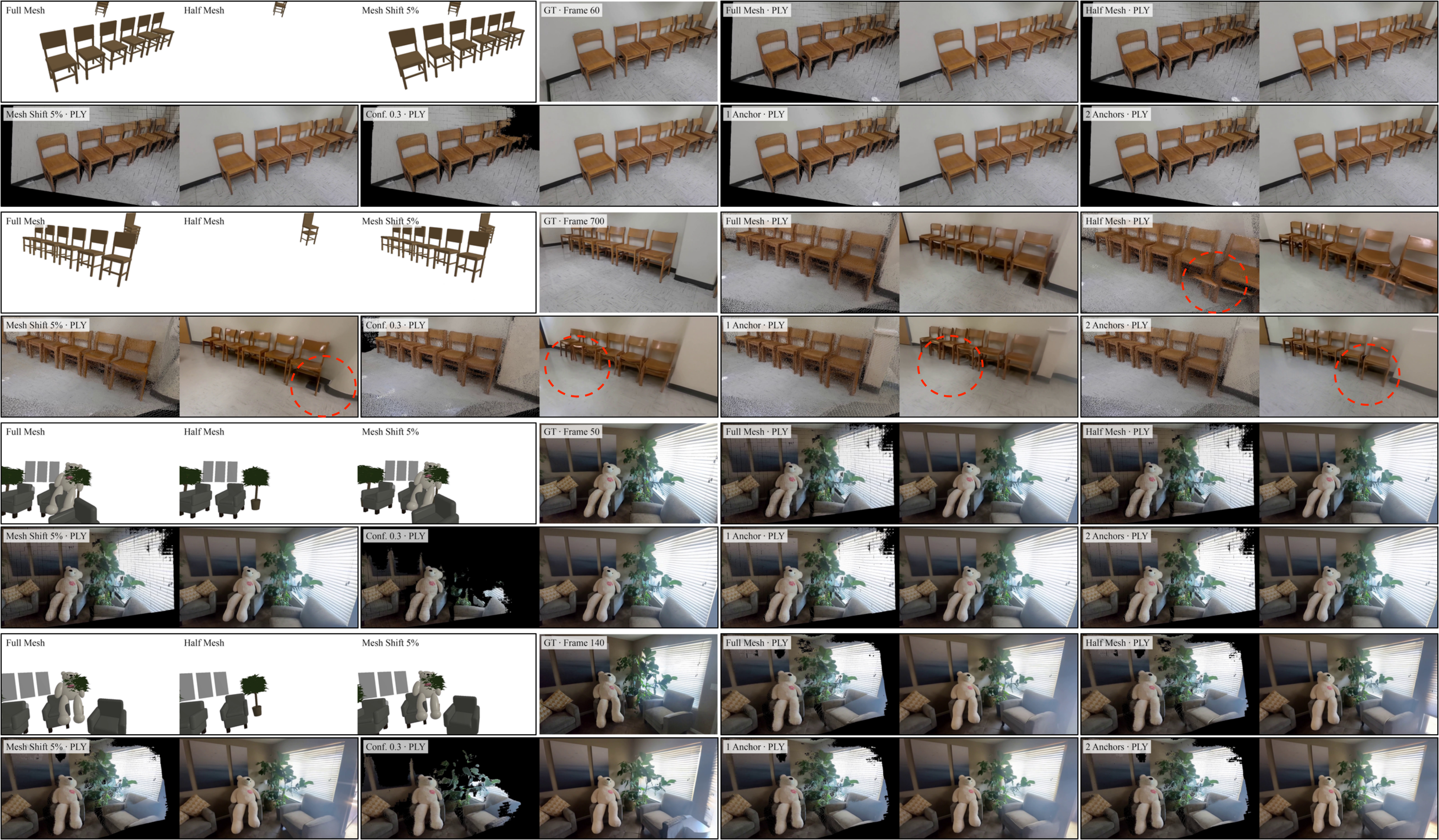}
\caption{\textbf{Robustness.} We compare full conditioning (including imperfect meshes, such as extra chairs or a misplaced bear) with missing meshes, mesh misalignment, incomplete point clouds, and fewer object anchors (left: point-cloud rendering; right: predicted frame). All variants generate videos with good overall visual quality. Accurate point-cloud guidance prevents incorrect background chair meshes from causing artifacts in the final video. Insufficient mesh guidance can distort fine structures such as chair legs, while incomplete point clouds reduce foreground and background detail. Fewer anchors can work well, as training includes scenes with only one or two anchors, but may cause object drift in multi-object scenes.}
\label{fig:upstream-conditioning-robustness}
\end{figure*}

\paragraph{Humanoid Setup and Application Details.}

Following GRAIL \cite{xie2026grail}, we first generate a reference human–object interaction (HOI) video of sitting, reconstruct the corresponding 4D HOI trajectory, and retarget the human motion to Unitree G1. We then apply GRAIL's scene-aware RL tracking method to assess whether the retargeted motion can be executed in the corresponding chair scene, fine-tuning a pretrained SONIC \cite{luo2026sonic} whole-body controller with PPO in Isaac Lab. The policy conditions on reference motions, proprioceptive and previous-action histories, chair-relative state, and a CNN-encoded local height map. The controller and scene encoder are fine-tuned together, with the chair represented as fixed collision geometry. Training rewards encourage reference tracking of root/body positions, orientations and velocities, together with five-point positional tracking at the torso, wrists and ankles. Smoothness, joint-limit and undesired-contact penalties regularize the behavior, while an auxiliary motion-reconstruction loss stabilizes training and early termination handles excessive tracking error. In Fig. 8, geometric parameters include object dimensions, surface curvature, component thickness, and structural details, such as wheel radii or chair-leg shapes. Spatial parameters control position, orientation, and scale, while appearance parameters include color, roughness, and metallic properties. An initial mesh provides geometric guidance for VivaWorld's video generation. The generated views provide additional visual guidance about silhouettes, object parts, and spatial relationships. Astra uses these views as visual references to revise the procedural mesh-generation code, refining proportions, surface details, component placement, and materials. The revised mesh is then rendered from multiple viewpoints to inspect its consistency before guiding another round of video generation.

\section{Experiments}
\label{app:long-video-style}
\paragraph{Camera Evaluation}For long-video style score, we compare consecutive anchors at per chunk's final frame, using the same anchors for every method. This measures local appearance stability and may miss gradual global drift, which differs from comparing only the first and last frames. Single-chunk and first-chunk style retain their endpoint comparisons within the evaluated windows. The reference trajectory is the ViPE trajectory of the ground-truth video. We use ViPE rather than the dataset-provided poses because it provides dense per-frame poses and intrinsics with one estimator for both DL3DV and Tanks and Temples. The trajectory of each generated window is estimated with DROID-SLAM, the camera pose estimator of WorldScore, using the reference per-frame intrinsics, so only extrinsics are estimated. Both trajectories are expressed relative to the first camera of the window. We first normalize the reference translations by the maximum displacement of
the reference camera from its first position within the window, so that translation errors are a fraction of the trajectory extent rather than a metric distance. Rotations are compared without further alignment, while the monocular scale ambiguity of the estimated trajectory is removed by a single positive global scale factor, chosen to minimize the sum of per-frame Euclidean distances between the scaled estimated and the normalized reference camera positions. In addition, Cam.\ applies clipping and nonlinear normalization to each scene's errors before averaging: rotation errors above $15^\circ$ and translation errors above $0.5$ are clipped to these thresholds, so further increases do not affect the score. RotErr is the geodesic angle between the estimated and reference camera rotations, measured in degrees, and TransErr is the Euclidean distance between the scaled estimated and the normalized reference camera position.

\paragraph{Upstream Error Analysis.}
As shown in Table~\ref{tab:upstream_sensitivity}, we construct failure cases on 5 randomly selected test scenes by retaining approximately half the meshes (reducing the average from 9 to 4 objects per scene, the mesh number in 5 cases ranges from 3 to 17), filtering out 44.7\% of the point-cloud (PLY) points with a confidence threshold of 0.3, and reducing the number of object anchors to 3, 2 and 1 to evaluate sensitivity to degraded upstream conditioning. Removing 44.7\% point-cloud points substantially reduces conditioning PSNR (around - 2.2) and slightly decreases generation PSNR (-0.20) , while shifting mesh positions with 5\% offset degrades performance a little as well. Using one or two anchors causes relatively small changes (since in our training dataset, some scenes only have 1-2 anchors), and retaining half the meshes has a modest impact. This tolerance to incomplete geometry is supported by our training augmentations and the varying mesh coverage provided by sparser RecGen meshes (might introduce some inaccurate mesh as well for the feed forward curation in the training dataset) and denser Astra meshes (might introduce some coarse mesh shape), using progressive point-cloud reconstruction from generative frames (might have inaccurate point cloud conditions) and full ground truth global point clouds.

\vspace{-5pt}
\paragraph{Runtime and Memory.}
On a single NVIDIA A100 80GB GPU, our method generates 1,000 frames at $832 \times 480$ resolution in 166.4 minutes, with a peak memory usage of 44.76~GiB. At the same resolution, our method is faster than VerseCrafter, Lyra2, NeoVerse, and Voyager, but slower than Gen3C and AnyRecon. Diffusion denoising accounts for 150.27 minutes, other processing across chunks takes 7.31 minutes, and preprocessing and remaining overhead take 8.82 minutes. Tables~\ref{tab:runtime_comparison} and~\ref{tab:runtime_breakdown} provide the runtime comparison and detailed breakdown, respectively.

\begin{table*}[htbp]
\centering
\caption{\textbf{Evaluation on 3D Scene Reconstruction with DAv3 and 3DGS optimization.}}
\label{tab:3dgs_optimization}
\normalsize
\setlength{\tabcolsep}{4pt}
\begin{tabular*}{\textwidth}{@{\extracolsep{\fill}}lcccccccc@{}}
\toprule
& \multicolumn{4}{c}{Short-Video}
& \multicolumn{4}{c}{Long-Video} \\
\cmidrule(lr){2-5}
\cmidrule(lr){6-9}
& \multicolumn{2}{c}{SSIM $\uparrow$}
& \multicolumn{2}{c}{LPIPS $\downarrow$}
& \multicolumn{2}{c}{SSIM $\uparrow$}
& \multicolumn{2}{c}{LPIPS $\downarrow$} \\
\cmidrule(lr){2-3}
\cmidrule(lr){4-5}
\cmidrule(lr){6-7}
\cmidrule(lr){8-9}
Method
& Before & After
& Before & After
& Before & After
& Before & After \\
\midrule
AnyRecon
& 0.461 & 0.496 & 0.449 & 0.462
& \cellcolor{ScoreSecond}{0.444} & \cellcolor{ScoreSecond}{0.431} & 0.673 & 0.633 \\
Gen3C
& \cellcolor{ScoreSecond}{0.472} & 0.544 & 0.448 & 0.446
& 0.418 & 0.394 & 0.676 & 0.637 \\
Voyager
& 0.461 & 0.504 & 0.492 & 0.492
& 0.350 & 0.276 & 0.802 & 0.828 \\
Lyra2
& \cellcolor{ScoreSecond}{0.472} & \cellcolor{ScoreSecond}{0.555} & \cellcolor{ScoreSecond}{0.439} & \cellcolor{ScoreSecond}{0.426}
& 0.423 & 0.406 & \cellcolor{ScoreSecond}{0.658} & \cellcolor{ScoreSecond}{0.616} \\
VerseCrafter
& 0.451 & 0.495 & 0.475 & 0.446
& 0.403 & 0.383 & 0.763 & 0.748 \\
NeoVerse
& 0.470 & 0.462 & 0.460 & 0.479
& 0.415 & 0.330 & 0.781 & 0.738 \\
\textbf{Ours}
& \cellcolor{Palette4}{\textbf{0.480}} & \cellcolor{Palette4}{\textbf{0.569}}
& \cellcolor{Palette4}{\textbf{0.417}} & \cellcolor{Palette4}{\textbf{0.406}}
& \cellcolor{Palette4}{\textbf{0.447}} & \cellcolor{Palette4}{\textbf{0.435}}
& \cellcolor{Palette4}{\textbf{0.598}} & \cellcolor{Palette4}{\textbf{0.568}} \\
\bottomrule
\end{tabular*}
\end{table*}

\vspace{-5pt}
\section{More Comparison Results and Showcases}
\vspace{-5pt}
\subsection{Comparison}
\vspace{-5pt}
\paragraph{3D Scene Evaluation}
Following Lyra2, we use Gaussians predicted by DA3 (Depth Anything 3). We additionally evaluate performance after 3DGS optimization, with results reported in Table~\ref{tab:3dgs_optimization}. 3DGS optimization generally improves short-video SSIM but reduces long-video SSIM, possibly because accumulated inconsistencies make long sequences harder to fit with a single static scene. Our method retains the best SSIM and LPIPS in both settings, before and after optimization.

\begin{figure*}[p]
\centering
\includegraphics[width=\textwidth]{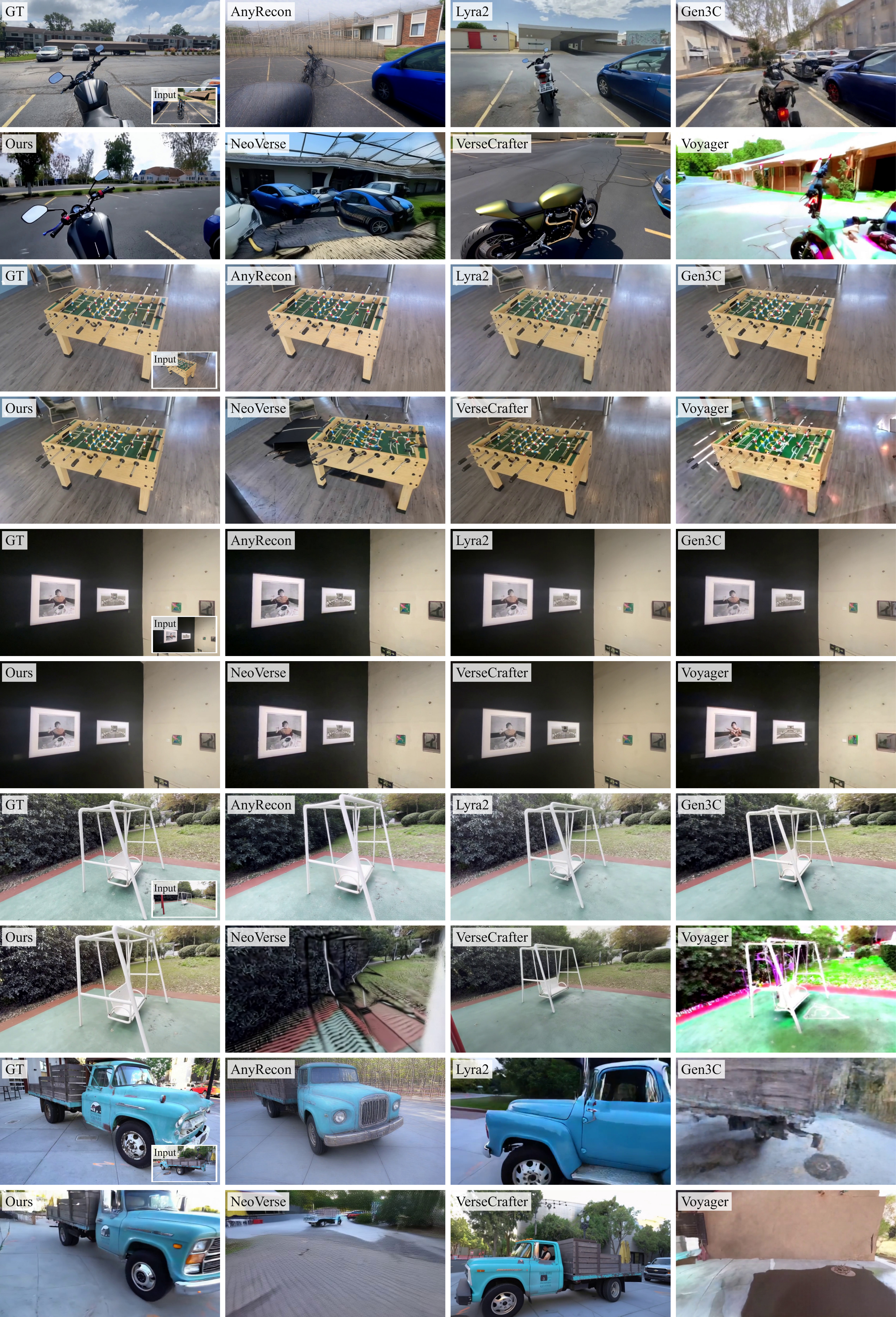}
\caption{\textbf{Additional Video Generation Results.}}
\label{fig:appendix-video-comparison}
\end{figure*}

\begin{figure*}[t]
\centering
\includegraphics[width=\textwidth]{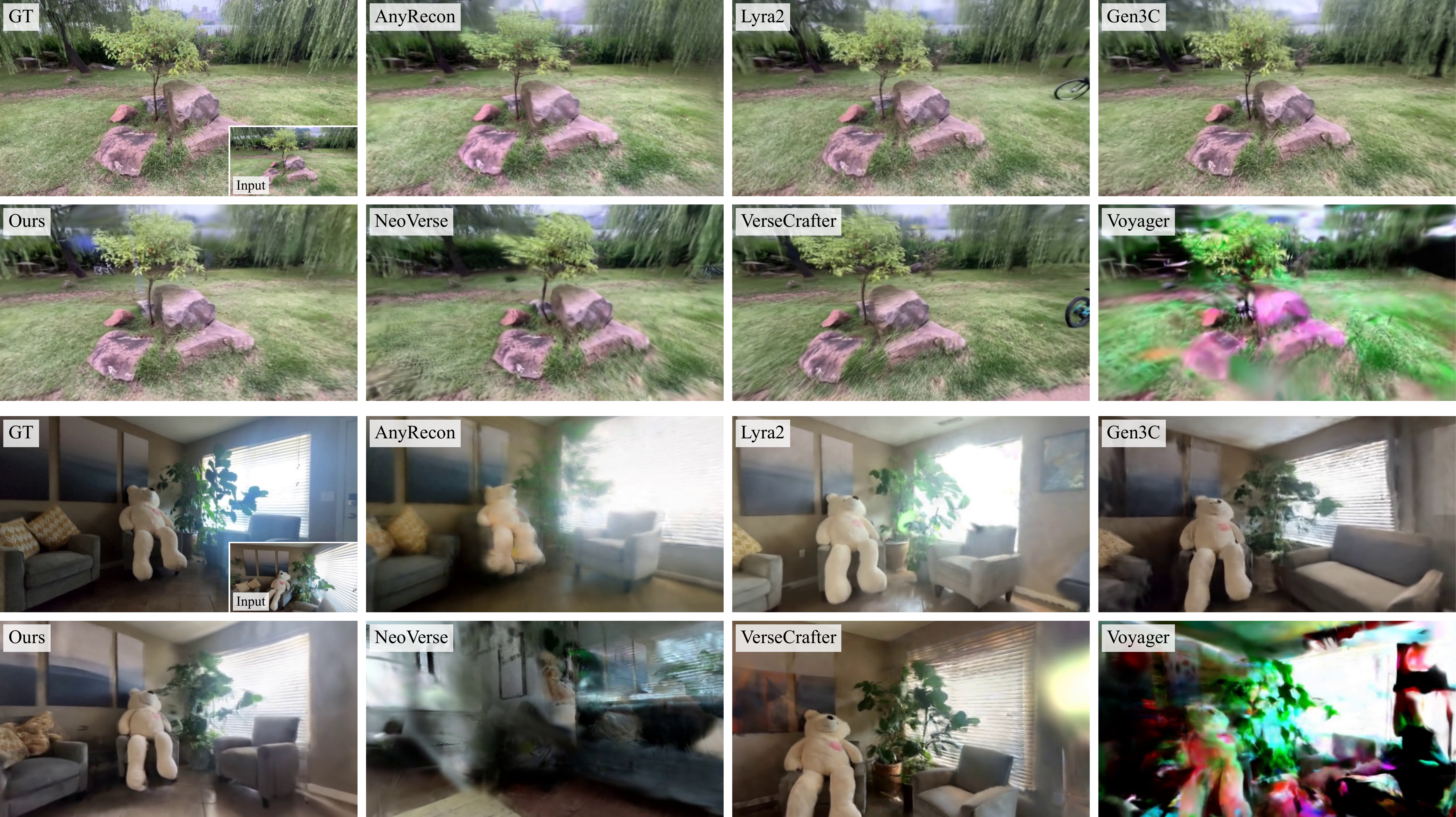}
\vspace{-10px}
\caption{\textbf{Additional 3D Scene Reconstruction Results.}}
\vspace{-2pt}
\label{fig:appendix-scene-reconstruction}
\end{figure*}

\textbf{Qualitative Comparison.}We provide additional qualitative comparisons of long video generation and 3DGS scene reconstruction in Fig.~\ref{fig:appendix-video-comparison} and Fig.~\ref{fig:appendix-scene-reconstruction}, following the same setting as in Sec. 4.3 (Qualitative Eval). Compared with the baselines, our method better preserves the object layout, appearance, and camera trajectory throughout long video generation. When the generated videos are used for 3DGS scene reconstruction, our method also yields cleaner and more coherent geometry, whereas the baselines suffer from blur, duplicated structures, and floating artifacts.

\begin{figure*}[!t]
\centering
\includegraphics[width=\textwidth]{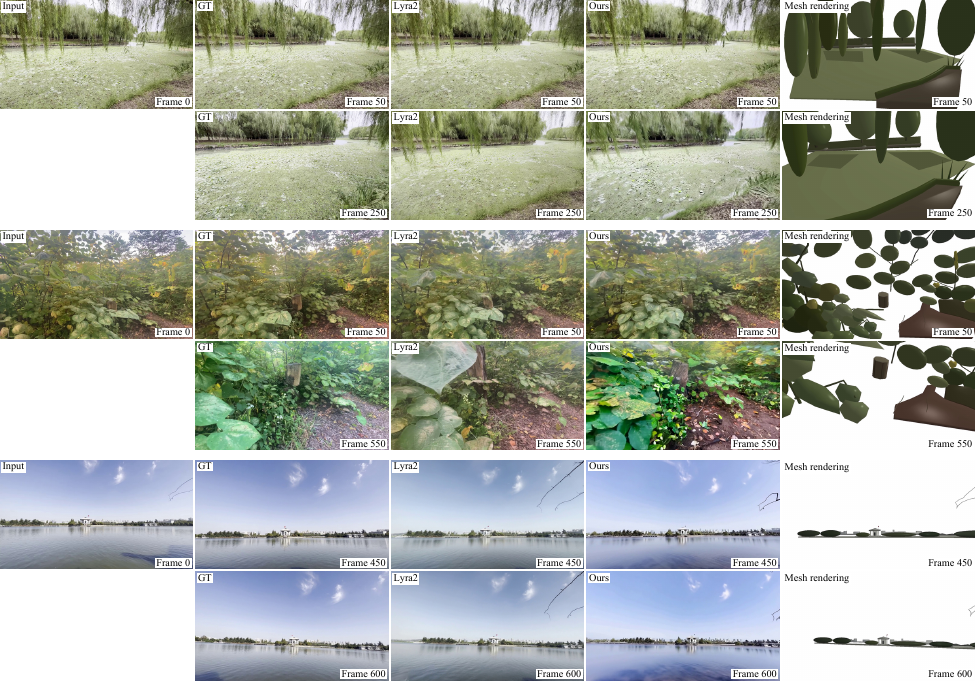}
\caption{\textbf{Additional Video Generation on Landscape Scenes.} From left to right: input, GT, Lyra2, Ours, and mesh conditioning. Mesh panels show the corresponding first-frame mesh guidance.}
\label{fig:appendix-e-landscape}
\end{figure*}

\begin{wraptable}{r}{0pt}
    \small
    \setlength{\tabcolsep}{3pt}
    \begin{tabular}{@{}lccccc@{}}
        \toprule
        Method & PSNR $\uparrow$ & SSIM $\uparrow$ & LPIPS $\downarrow$ & Cam. $\uparrow$ & Style $\uparrow$ \\
        \midrule
        Ours & \textbf{13.69} & \textbf{0.3797} & \textbf{0.5452} & \textbf{70.24} & \textbf{97.41} \\
        Lyra2 & 13.17 & 0.3770 & 0.6534 & 58.46 & 89.60 \\
        \bottomrule
    \end{tabular}
\end{wraptable}

\begin{figure*}[htbp]
\centering
\includegraphics[width=\textwidth]{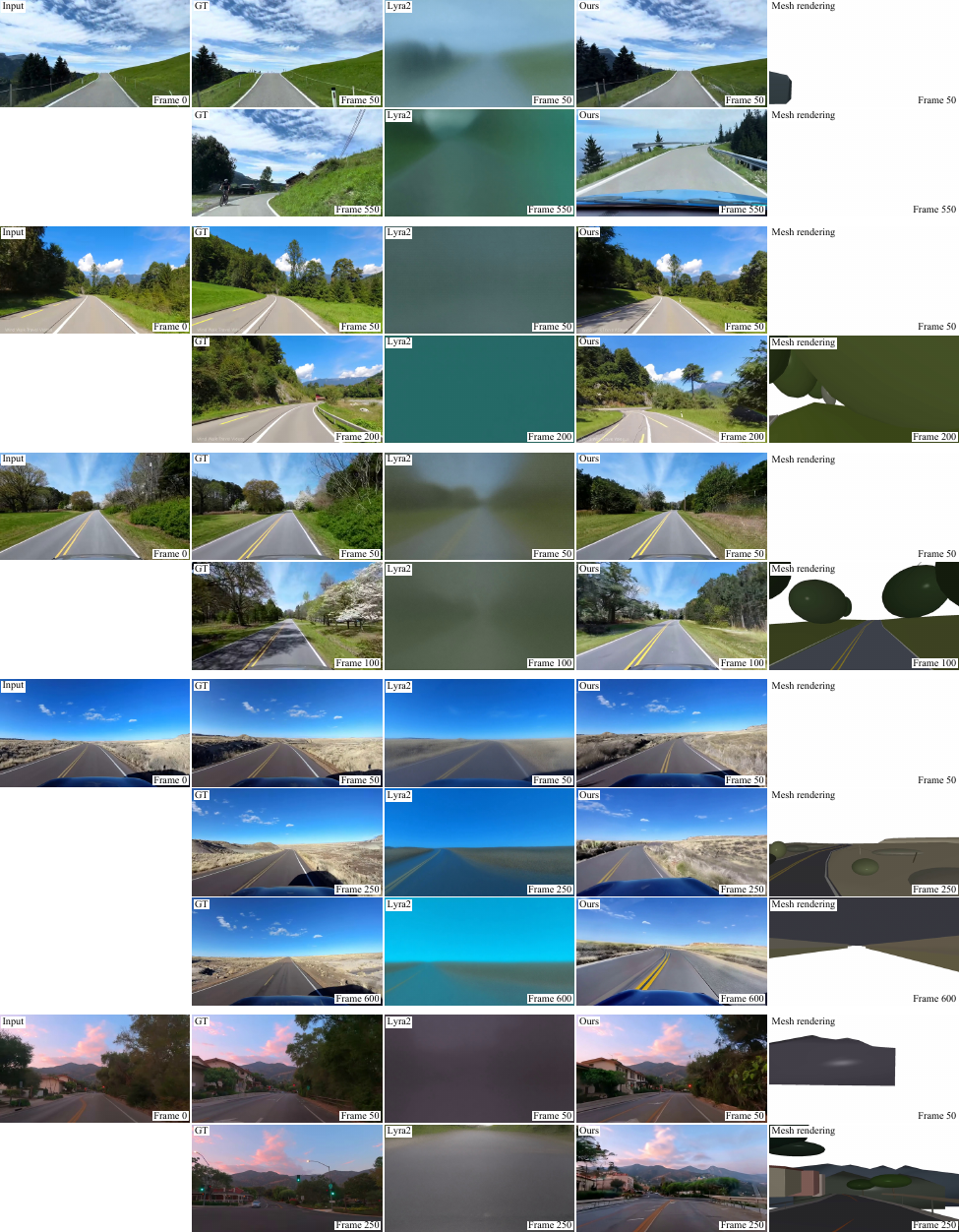}
\caption{\textbf{Additional Video Generation on Driving Scenes.} From left to right: input, GT, Lyra2, Ours, and prediction-derived mesh conditioning. Mesh panels show the corresponding per-chunk Astra guidance.}
\label{fig:appendix-e-driving}
\end{figure*}

\vspace{-8pt}
\paragraph{Comparison on Landscape and Autonomous Driving (AV) Scenes} To further demonstrate our method's performance with unreliable mesh conditioning, we select five natural scenes from DL3DV, including lakes and mountains, where we only have coarse meshes with incorrect shapes or placements. We use these imperfect Astra meshes during inference. We also evaluate five more challenging autonomous-driving scenes from OpenDV \cite{yang2024generalized}, which is outside the DL3DV dataset used to train both our method and Lyra2 and we have no mesh rendering in some cases. For these driving scenes, we use Astra to update the mesh conditioning per chunk as the camera moves forward. We compare with the strongest baseline, Lyra2, on all ten scenes. The right table shows that our method achieves better image quality, camera control, and style consistency, even with unreliable meshes featuring coarse shapes or incorrect placements. We provide selected additional qualitative comparisons on landscape and driving scenes in Figs.~\ref{fig:appendix-e-landscape} and~\ref{fig:appendix-e-driving}. The landscape examples use fixed initial Astra meshes and the driving examples show prediction-derived per-chunk Astra mesh updates.

\clearpage
\clearpage

\subsection{More Showcases} 

We provide additional results on long video generation and 3DGS reconstruction following the same setting in the qualitative study. Our method achieves accurate camera control and strong geometric consistency in long video generation. Moreover, when the generated long videos are used for 3D reconstruction, our method still maintains strong consistency in geometry, shape, and camera trajectory.

\begin{figure*}[htbp]
\centering
\includegraphics[width=\textwidth]{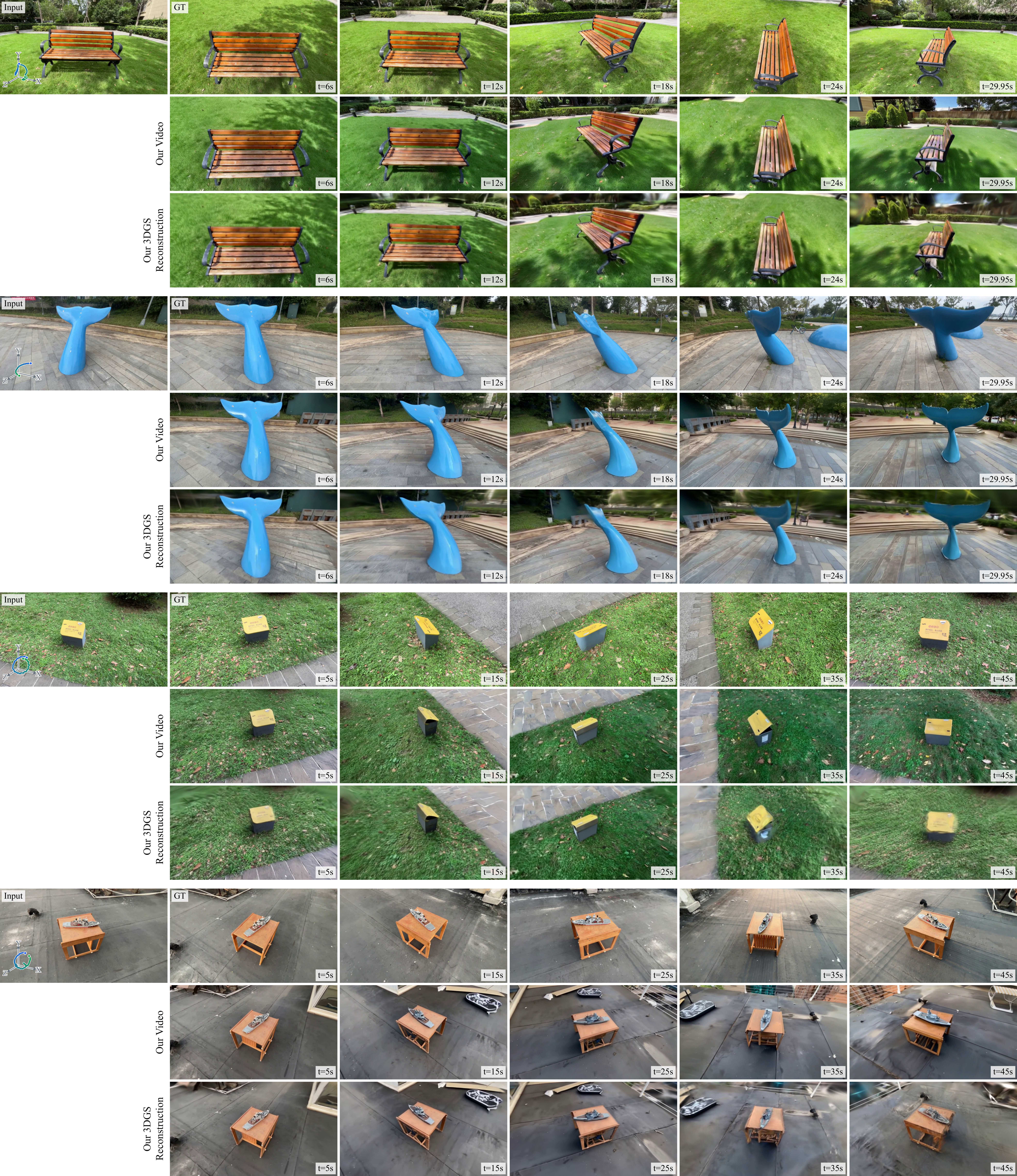}
\caption{\textbf{Additional Video Generation and 3D Scenes.} For each case: input image and ground-truth frames (top), our generated video (middle), and our 3DGS reconstruction rendered along the same trajectory (bottom). Numbers denote seconds.}
\label{fig:appendix-results-11}
\end{figure*}

\begin{figure*}[htbp]
\centering
\includegraphics[width=\textwidth]{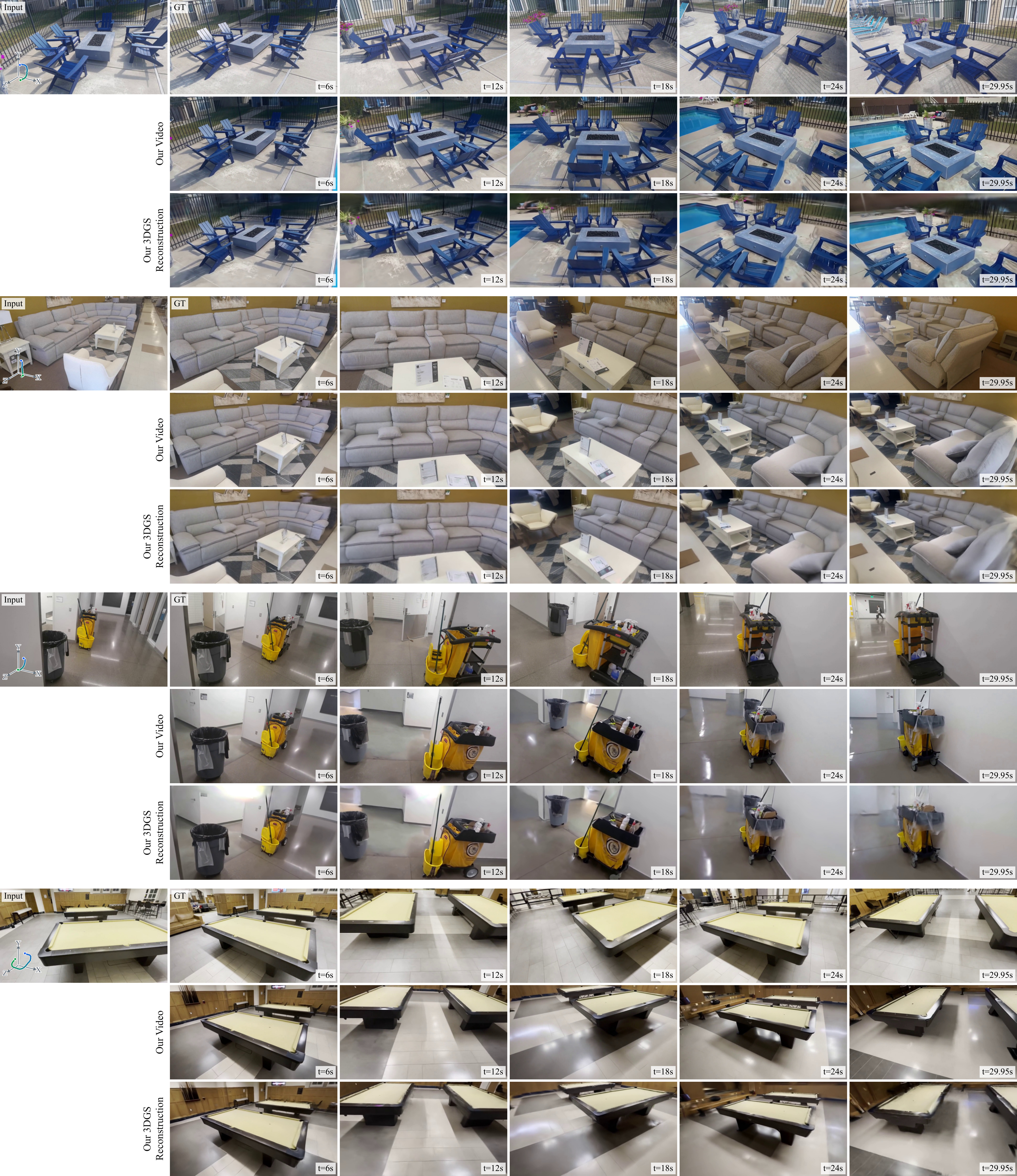}
\caption{\textbf{Additional Video Generation and 3D Scenes.} For each case: input image and ground-truth frames (top), our generated video (middle), and our 3DGS reconstruction rendered along the same trajectory (bottom). Numbers denote seconds.}
\label{fig:appendix-results-12}
\end{figure*}

\clearpage

\begin{figure*}[htbp]
\centering
\includegraphics[width=0.95\textwidth,height=0.85\textheight,keepaspectratio]{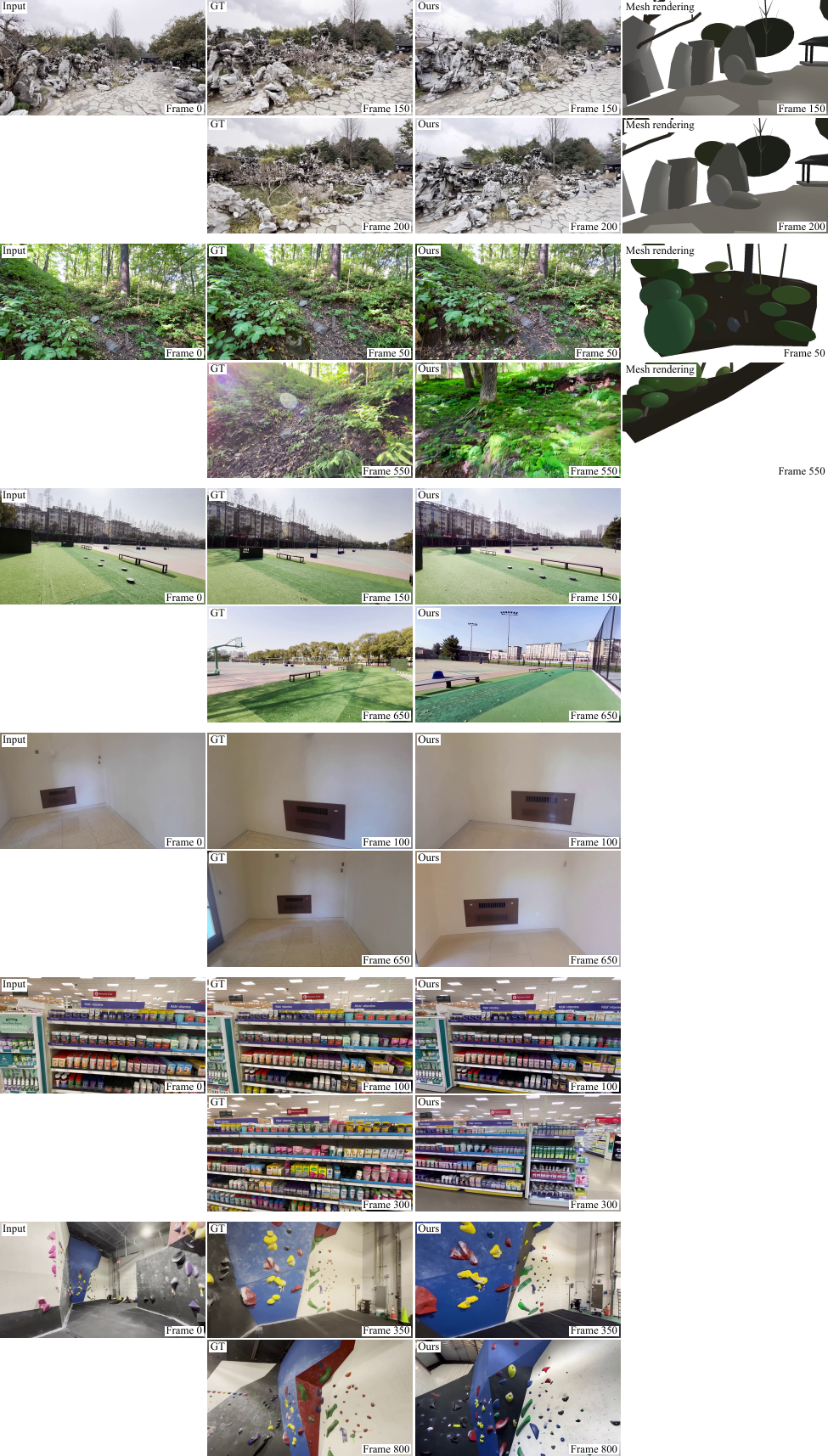}
\caption{\textbf{Additional Video Generation.} From left to right: input, GT, Ours, and mesh conditioning. Blank mesh cells in the last four scenes indicate only point-cloud guidance. }
\label{fig:appendix-e-additional-showcases}
\end{figure*}

To further demonstrate robustness to unusable meshes, we show two DL3DV scenes where RecGen fails, using only Astra meshes with incorrect shapes and placements (top two cases) and four cases using only progressively reconstructed point clouds and global camera conditioning. In Fig.~\ref{fig:appendix-e-additional-showcases}, our model still generates realistic videos, although quality may be lower than with the full pipeline.

\end{document}